\documentclass{article} 
\usepackage{iclr2027_conference,times}

\usepackage{amsmath,amsfonts,bm}

\def\eqref#1{equation~\ref{#1}}

\def\1{\bm{1}}

\DeclareMathAlphabet{\mathsfit}{\encodingdefault}{\sfdefault}{m}{sl}
\SetMathAlphabet{\mathsfit}{bold}{\encodingdefault}{\sfdefault}{bx}{n}

\usepackage{amsmath,amssymb}
\usepackage{graphicx}
\usepackage{tabularx}
\usepackage{tabu}
\usepackage{tabularray}
\usepackage{multirow}
\usepackage{booktabs}
\usepackage{colortbl}
\usepackage{capt-of}
\usepackage{wrapfig}
\usepackage[section]{placeins}
\usepackage{microtype}
\usepackage{hyperref}
\definecolor{citationblue}{HTML}{315F86}
\hypersetup{
  colorlinks=true,
  citecolor=citationblue,
  linkcolor=black,
  urlcolor=black
}
\usepackage{url}

\newcommand{\OURS}{\textsc{MedEVM}}
\newcommand{\METHOD}{\textsc{EVD-Harness}}

\title{Do LLMs Value the Right Evidence? Evidence--Value \mbox{Misalignment} in Dynamic Medical Diagnosis}

\author{Kehua Feng$^{1,2,*}$, Yunsheng Lu$^{1,2}$, Yitong Qiao$^{1,2}$, Tiantian He$^{2}$, Lei Liu$^{1,2,\dagger}$, Yue Shen$^{2}$\\ 
{\normalsize\bfseries Jian Wang$^{2}$, Jinjie Gu$^{2}$} \\
  $^1$Zhejiang University \;
  $^2$Ant Healthcare, Ant Group \\
\texttt{kehuafeng@zju.edu.cn}; \texttt{liulei1497@gmail.com} \\
}

\iclrfinalcopy
\begin{document}
\renewcommand{\thefootnote}{}
\footnotetext[0]{$^*$Work done during Kehua's internship at Ant Group.}
\footnotetext[1]{$^\dagger$Corresponding Author.}
\maketitle
\lhead{\parbox{\textwidth}{\footnotesize Do LLMs Value the Right Evidence? Evidence--Value Misalignment in Dynamic Medical Diagnosis}}
\setlength{\headheight}{22pt}

\begin{abstract}

A correct diagnosis reached from insufficient or misleading evidence can pose a clinical hazard, yet outcome-based accuracy may reward such lucky guesses. We call this mismatch between diagnostic decisions and the value of available evidence \textbf{Evidence-Value Misalignment (\textsc{EVM})}. To disentangle evidential grounding independently from diagnostic accuracy, we introduce \OURS{}, a dynamic benchmarking environment comprising 1,050 cases across 24 disease systems. Observations arrive turn by turn, requiring models to continuously calibrate its decision by deciding whether to wait for more evidence or submit a diagnosis. Across 9 LLMs, four interesting patterns are observed. (1) \textbf{Miscalibrated evidence tracking}. Making a diagnosis often fails to calibrate evidence sufficiency, even in more capable models, and even worsens in reasoning mode. (2) \textbf{Misaligned diagnosis submission}. Confidence in the correct diagnosis often fails to ensure timely submission despite sufficient evidence. (3) \textbf{Evidence order matters}. Reordering the same evidence changes diagnoses even when model confidence remains similar. (4) Misleading evidence remains influential. Added misleading evidence redirects diagnoses even after prior evidence becomes sufficient. We further verify that EVM predicts errors and that preventing premature submission improves accuracy. These findings motivate \textbf{Evidence-Verified Diagnosis Harness (\METHOD{})}. It decouples diagnosis generation from submission through an offline Contrastive Diagnostic Wiki and three online control stages, namely observation management, proposal and witness verification, and diagnosis submission control. Across five LLMs, \METHOD{} improves accuracy by 12.0--51.1 percentage points while mitigating EVM-related failures. Our results demonstrate that verifying evidential support before submission can make diagnostic decisions more reliable.

\end{abstract}

\vspace{-0.6em}\section{Introduction}
\label{sec:introduction}

Large language models (LLMs) show strong reasoning capabilities on difficult medical questions and complex cases \citep{singhal2023clinical}. To evaluate these capabilities beyond static medical QA \citep{jin2021medqa,pal2022medmcqa,zhu2026diagnosisarena}, AgentClinic~\citep{schmidgall2024agentclinic}, AI Hospital~\citep{fan2025aihospital}, and MediQ~\citep{li2024mediq} introduce multi-turn diagnostic interactions in which models gather patient information and refine diagnoses, bringing evaluation closer to real clinical encounters. EHRWorld~\citep{mu2026ehrworld} and ChronoMedicalWorld~\citep{wang2026chronomedicalworld} further extend these interactive settings by modeling how patient states evolve under clinical interventions.

\begin{figure}[!ht]
  \centering
  \vspace{-1.5em}
  \includegraphics[width=\linewidth]{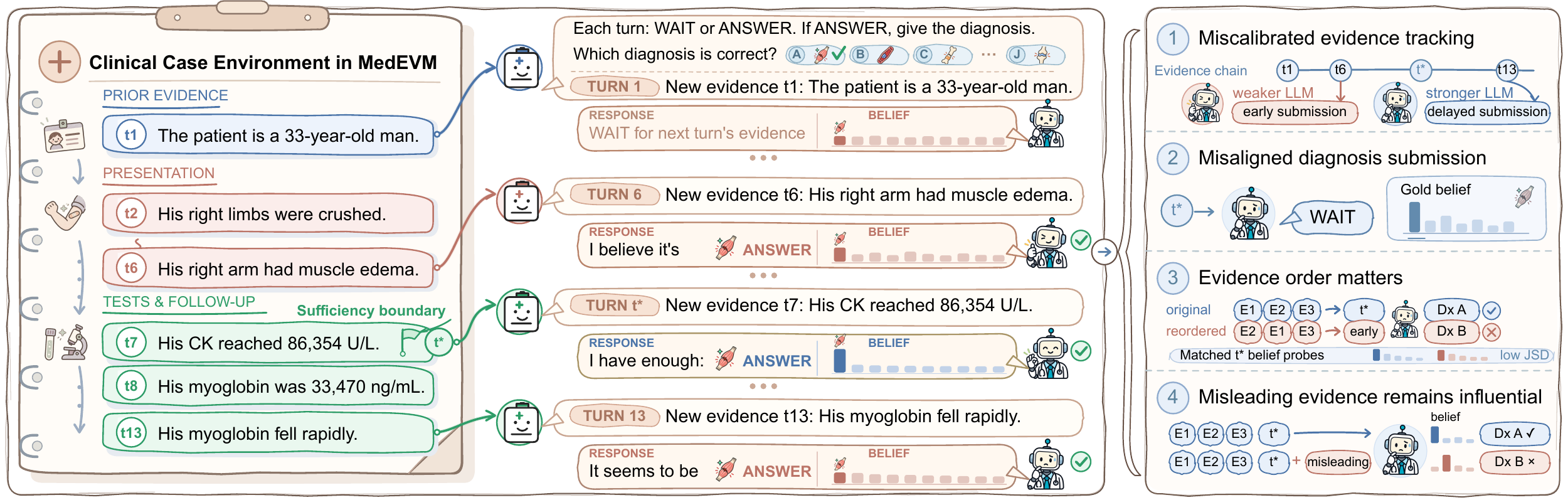}\vspace{-1em}
  \caption{Overview of \OURS{}. Final accuracy alone obscures differences in when models submit diagnoses. In \OURS{}, clinical evidence arrives turn by turn, and the LLM Agent must wait or submit a diagnosis. This evaluation reveals four key findings hidden by answer-only scoring.}
  \label{fig:medevm-overview}
  \vspace{-0.5em}
\end{figure}



Yet increasingly realistic clinical environments do not make final diagnostic accuracy a sufficient measure of clinical reasoning. Accuracy-based evaluation treats a correct diagnosis submitted on insufficient evidence the same as one supported by sufficient evidence, leaving the evidential basis for submission unexamined. Although recent work has examined vulnerabilities in multi-turn settings and premature diagnosis submission \citep{laban2026lost,fang2026mint,hsu2026medaction}, a fundamental question remains: Do LLMs appropriately weigh the available evidence when deciding to submit a diagnosis? Here, evidence value refers to an observation’s ability to narrow the differential diagnosis and inform whether the available evidence warrants submission.

To answer this question, we introduce \OURS{}, a progressive-evidence diagnosis benchmark comprising 1,050 cases across 24 disease systems (Section~\ref{sec:benchmark}). Each case consists of an ordered sequence of atomic observations extracted from a case report, a gold diagnosis, and nine plausible alternative diagnoses. After each observation, the model must decide whether to wait for more evidence or submit a diagnosis. We annotate the sufficiency boundary $t^*$ as the earliest turn at which the accumulated evidence is sufficient to distinguish the gold diagnosis from all alternatives. We define Evidence-Value Misalignment (EVM) as a mismatch between when a model submits a diagnosis and when the available evidence becomes sufficient to support that submission.
  

Controlled experiments across nine LLMs reveal four EVM patterns (Section~\ref{sec:mechanism}). (1) \textbf{Miscalibrated evidence tracking}. Making a diagnosis often fails to calibrate evidence sufficiency, even in more capable models, and even worsens in reasoning mode. (2) \textbf{Misaligned diagnosis submission}. Confidence in the correct diagnosis often fails to ensure timely submission despite sufficient evidence. (3) \textbf{Evidence order matters.} Diagnostic decisions are sensitive to evidence order, even when model confidence remains similar. (4) Misleading evidence remains influential. Added misleading observations can redirect diagnoses regardless of whether prior evidence was sufficient.


We next examine whether EVM helps explain diagnostic decisions and errors (Section~\ref{sec:impact}). Measures of evidence valuation provide information beyond model confidence for predicting when models submit a diagnosis and whether that diagnosis is correct. We then test the effect of correcting premature submission using \textit{Natural Gate}, an oracle intervention that defers submission until the annotated sufficiency boundary $t^*$. Among cases in which models originally submitted before $t^*$, this intervention improves accuracy by 29.6--48.4 percentage points.


These findings motivate the Evidence-Verified Diagnosis Harness (\METHOD{}), which separates diagnosis generation from submission using an offline Contrastive Diagnostic Wiki and three online control stages (Section~\ref{sec:quorum}). Observation management organizes incoming evidence and retrieves relevant diagnostic knowledge. Proposal and witness verification generates candidate diagnoses and checks the evidence supporting them. Diagnosis submission control validates the cited evidence before allowing a diagnosis to be submitted. Across five LLMs, \METHOD{} improves diagnostic accuracy by 12.0--51.1 percentage points while reducing EVM-related failures. These results show why valuing the right evidence should be evaluated separately from producing the right answer. Our contributions are threefold:



\begin{itemize}
  \item We introduce \OURS{}, a progressive-evidence benchmark that evaluates diagnosis submission against annotated evidence sufficiency, revealing failures that final diagnostic accuracy alone cannot capture.

  \item We identify Evidence-Value Misalignment (EVM) through controlled analyses of submission timing and responses to reordered and misleading evidence. We show that EVM measures help identify diagnostic errors, while delaying premature submission improves accuracy.

  \item We develop \METHOD{}, which separates diagnosis generation from submission control, better aligning submission with evidence sufficiency and improving diagnostic accuracy.
\end{itemize}

\section{Related Work}
\label{sec:related-work}

\textbf{Dynamic Medical Diagnosis Benchmarks.}
Static benchmarks such as MedQA~\citep{jin2021medqa}, MedMCQA~\citep{pal2022medmcqa}, and PubMedQA~\citep{jin2019pubmedqa} evaluate medical knowledge and final answers from fully presented contexts, but cannot reveal how diagnosis evolves as evidence arrives. Recent benchmarks instead model diagnosis as a dynamic process. DDXPlus~\citep{fansi2022ddxplus} supports symptom-seeking differential diagnosis, while AgentClinic~\citep{schmidgall2024agentclinic}, AI Hospital~\citep{fan2025aihospital}, and MMD-Eval~\citep{liu2025mmdeval} simulate consultations that involve patients, examinations, or tools. MediQ~\citep{li2024mediq} evaluates proactive acquisition of missing information. MINT~\citep{fang2026mint} reveals fixed evidence over turns to study rushing, lures, and self-correction, while MedEinst~\citep{chen2026medeinst} uses counterfactual cases to expose shortcut-driven errors hidden by endpoint accuracy. Complementary to these efforts, \OURS{} treats neither dialogue quality nor the final answer as the sole endpoint. It aligns turn-level belief and natural wait-or-answer behavior with an externally annotated earliest sufficiency boundary, then uses controlled interventions to test how misalignment changes diagnostic outcomes.

\textbf{From Uncertainty Estimation to Diagnosis Submission Control.}
Prior work estimates answer reliability through model self-evaluation, black-box confidence elicitation, and semantic uncertainty \citep{kadavath2022know,xiong2024uncertainty,farquhar2024semantic}. Selective prediction converts such scores into answer-or-abstain decisions \citep{kamath2020selective}, and MedAbstain~\citep{machcha2026abstain} extends this framework to medical MCQA under missing or perturbed context. For sequential agents, KnowNo~\citep{ren2023knowno} forms conformal action sets and requests help when the next action is not uniquely supported. Closest to our failure setting, \citet{handler2026premature} evaluate premature closure as inappropriate diagnosis submission under uncertainty. These approaches provide important reliability tools, but answer confidence is not equivalent to clinical evidence sufficiency. A model may submit a diagnosis early with high confidence or keep waiting after it becomes uniquely identifiable. We therefore treat diagnosis submission as an evidence-conditioned decision distinct from answer generation. This framing motivates mitigation that reduces premature submission without merely replacing it with delayed answers or persistent abstention.

\setcounter{section}{2}
\section{Dynamic \OURS{} Benchmark}
\label{sec:benchmark}

To systematically evaluate whether LLMs can align diagnosis submission with incrementally available evidence, we introduce \OURS, a progressive-evidence diagnosis benchmark that simulates a dynamic clinical environment by revealing atomic observations from case reports in sequential turns, and records when LLMs wait, submit a diagnosis, or revise their diagnostic beliefs.

\paragraph{Task Formulation.} We formulate progressive-evidence diagnosis as a sequential decision-making problem under partial observability. A clinical case $i$ comprises a hidden ground-truth diagnosis $y_i^*$, a candidate set $\mathcal{Y}_i$, and a fixed sequence of atomic evidence $E_i=(e_{i1},\ldots,e_{iT_i})$. At turn $t$, the environment reveals one new observation $e_{it}$ and appends it to the history $h_{it}=E_{i,\leq t}$. The LLM agent then samples an action $a_{it}\sim\pi_\theta(\cdot\mid h_{it})$ from $\mathcal{A}_i=\{\textsc{wait}\}\cup\{\textsc{answer}(y):y\in\mathcal{Y}_i\}$. Taking $\textsc{wait}$ advances the environment to $t+1$, while outputting an $\textsc{answer}$ irreversibly terminates the trajectory. Reaching $T_i$ without answering is considered a failure. Since the evidence schedule is fixed, the action determines when the trajectory ends but not which observation appears next.

\paragraph{Evidence-Value Misalignment.}

Let $\mathcal{C}_i(t)\subseteq\mathcal{Y}_i$ denote the subset of candidate diagnoses logically compatible with the history $h_{it}$. We define the \textbf{Sufficiency Boundary} as $ t_i^*=\min\{t:\mathcal{C}_i(t)=\{y_i^*\}\}.$

Before $t_i^*$, a confounder remains plausible. At $t_i^*$, the gold first becomes uniquely identifiable. We use its annotated position as the reference and test timing tolerance in Appendix~\ref{app:boundary-tolerance}. Let $s_{it}=\mathbf{1}[t\geq t_i^*]$ indicate whether the evidence is sufficient, and $c_{it}=\mathbf{1}[a_{it}=\textsc{answer}]$ denote whether the model submits a diagnosis. We decompose EVM into two mutually exclusive errors
\begin{equation}
m_{it}^{\mathrm{pre}}=c_{it}(1-s_{it}),\quad m_{it}^{\mathrm{delay}}=(1-c_{it})s_{it},\quad m_{it}=m_{it}^{\mathrm{pre}}+m_{it}^{\mathrm{delay}}=|s_{it}-c_{it}|.
\label{eq:evm-definition}
\end{equation}
Here, $m_{it}^{\mathrm{pre}}$ and $m_{it}^{\mathrm{delay}}$ measure premature and delayed diagnosis submission, respectively. EVM quantifies the alignment between decision timing and evidence sufficiency, which is distinct from diagnostic accuracy. A correct guess before $t_i^*$ is heavily penalized as a premature EVM error, while an incorrect answer after \(t_i^*\) is treated as a diagnostic error rather than a timing mismatch.

\paragraph{Benchmark Construction.} We collect case reports from PubMed-indexed journals \citep{white2020pubmed}, covering NEJM, BMJ Case Reports, Gastroenterology, JAMA, and The Lancet. After applying formatting and completeness filters, we retain 36,020 reports that contain a confirmed diagnosis and sufficient clinical information. We next remove diagnostic leakage, segment each narrative into typed atomic evidence in chronological order, and treat the confirmed diagnosis as ground truth. Using a structured prompt, GPT-5.4~\citep{openai2026gpt54} generates nine clinically plausible confounders for each case. Three proprietary models from different families, namely Claude Opus 4.8~\citep{anthropic2026claude48}, Gemini 3.1 Pro Preview~\citep{gemini31pro}, and GPT-5.6-sol~\citep{openai2026gpt56}, independently annotate the sufficiency boundary $t_i^*$, and we keep only cases with unanimous boundary annotations. Finally, Gemini 3.5 Flash~\citep{gemini35flash} and five graduate students with medical backgrounds jointly verify the quality of each sample, resulting in 1,050 cases for the final \OURS{} benchmark. Details can be found in Appendix~\ref{app:benchmark-construction}.

\begin{figure}[!ht]
  \centering
  \begin{minipage}[t]{0.60\textwidth}
    \vspace{0pt}
    \centering
    \includegraphics[width=\linewidth]{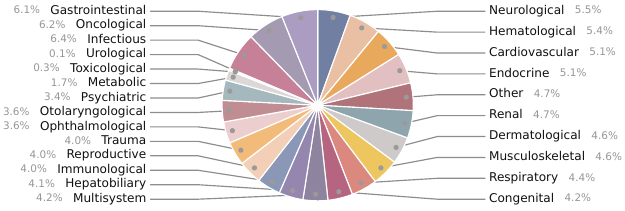}
    \setlength{\abovecaptionskip}{1.5pt}
    {\footnotesize\captionof{figure}{Distribution of all 1,050 \OURS{} benchmark cases across 24 disease systems.}}
    \label{fig:benchmark-construction}
  \end{minipage}\hfill
  \begin{minipage}[t]{0.38\textwidth}
    \vspace{0pt}
    \centering
    \scriptsize
    \vspace{-1em}\captionof{table}{Data statistics in different construction phases and final \OURS{}.}\vspace{1em}
    \renewcommand{\arraystretch}{1.15}
    \setlength{\tabcolsep}{2.1pt}
    \resizebox{0.9\linewidth}{!}{%
    \begin{tabular}{lrlr}
      \toprule
      \multicolumn{2}{c}{Construction Phases} & \multicolumn{2}{c}{Final \OURS{}} \\
      \cmidrule(lr){1-2}\cmidrule(lr){3-4}
      Phase & \#$N$ & Statistic & Value \\
      \midrule
      Collected & 75,434 & Observations & 26,869 \\
      Structured & 52,147 & Disease systems & 24 \\
      Qualified & 36,020 & Diagnoses & 953 \\
      Valid MCQ & 1,810 & Candidates/case & 10 \\
      \OURS{} & 1,050 & Evidence/case & 25.6 \\
      \bottomrule
    \end{tabular}
    }
    \label{tab:benchmark-statistics}
  \end{minipage}
\end{figure}

\textbf{Evaluation Protocol.} We evaluate 9 advanced LLMs, including the Qwen3 series~\citep{yang2025qwen3} with 8B (Q3-8B), 14B (Q3-14B), 32B (Q3-32B), and 235B-A22B (Q3-235B) parameters, Qwen3.6-27B~(Q3.6-27B; \citeauthor{qwen36_27}, \citeyear{qwen36_27}) and Qwen3.8-27B~(Q3.8-27B; \citeauthor{qwen38_27}, \citeyear{qwen38_27}) as similar-capacity variants, and DeepSeek-V4-Flash~(DS-V4-F; \citeauthor{xu2026deepseek}, \citeyear{xu2026deepseek}), Kimi-K2.5 (K2.5) and Kimi-K2.6~(K2.6; \citeauthor{kimiteam2026k25}, \citeyear{kimiteam2026k25}) from external families.
During evaluation, LLMs are constrained to output structured JSON responses. To measure the internal belief distribution without altering the natural trajectory, we apply a non-intervening probability probe to a cloned context. Given a forced JSON prefix up to the answer field, the probe extracts the exact sequence log-probability $\ell_{it}(y)$ of every candidate $y\in\mathcal{Y}_i$ and normalizes it as $p_{it}(y)=\exp(\ell_{it}(y))/\sum_{y'\in\mathcal{Y}_i}\exp(\ell_{it}(y'))$.

We report three recurring metrics: (1) \textit{Diagnosis Submission Accuracy}, correct diagnoses among answered trajectories, (2) \textit{Premature Submission Rate}, submissions before $t_i^*$ among all trajectories, and (3) \textit{Gold Belief} $p_{\mathrm{gold},it}:=p_{it}(y_i^*)$, defined as the normalized probability $p_{it}(y_i^*)$ of the correct answer. Experiment-specific metrics are introduced locally and defined in Appendix~\ref{app:metric-definitions}. Paired effects use case-clustered 95\% bootstrap confidence intervals unless stated otherwise.

\vspace{-0.5em}\section{How EVM Arises: Evidence and Diagnosis Submission Diverge}
\label{sec:mechanism}

\definecolor{evmtakeawaybg}{RGB}{247,247,255}
\definecolor{evmtakeawayline}{RGB}{171,176,232}
\providecommand{\evmtakeaway}[1]{%
  \par\vspace{0.1em}\noindent
  \begingroup\setlength{\fboxsep}{3.5pt}\setlength{\fboxrule}{0.45pt}%
  \fcolorbox{evmtakeawayline}{evmtakeawaybg}{\parbox{\dimexpr\linewidth-2\fboxsep-2\fboxrule\relax}{\footnotesize \textsc{\textbf{Takeaway}}: #1}}%
  \endgroup\par\vspace{0.1em}}

\begingroup
\setlength{\parskip}{4pt}
\setlength{\intextsep}{8pt plus 2pt minus 2pt}
\setlength{\textfloatsep}{10pt plus 2pt minus 2pt}

\begin{wrapfigure}[13]{r}{0.42\linewidth}
  \vspace{-1.0em}
  \centering
  \includegraphics[width=\linewidth,trim=0 4pt 0 0,clip]{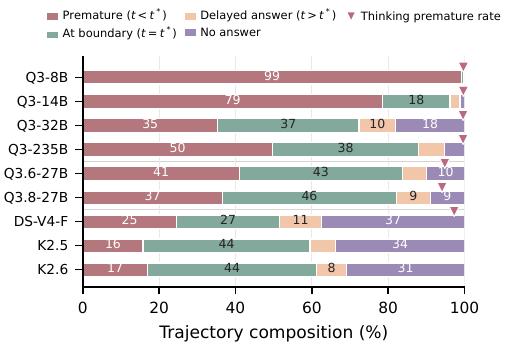}\vspace{-0.5em}
  \caption{Case proportions by diagnosis submission time relative to the first sufficient-evidence turn $t^*$. Red triangles show early submission with thinking.}
  \label{fig:evm-phenotypes}
\end{wrapfigure}

We examine EVM in medical diagnosis through four progressive experiments on its prevalence and mechanisms. These characterize diagnosis submission timing, probe diagnostic knowledge, test evidence order, and vary misleading evidence strength.


\paragraph{RQ1: Does capability reduce EVM?}\label{rq:timing}
We categorize trajectories by whether the model submits a diagnosis before, at, or after the sufficiency boundary $t^*$, or remains unanswered after all evidence arrives. Figure~\ref{fig:evm-phenotypes} compares model sizes, similar-capacity versions, and model families. We also enable thinking mode in seven models to test whether additional reasoning improves diagnosis submission timing.

All nine models exhibit substantial EVM, with capability gains changing its form rather than steadily reducing it. Qwen3-8B submits early in 99.1\% of cases, whereas larger and later LLMs reduce early submission but often wait despite sufficient evidence or never answer. These patterns persist when submissions within one or two turns of the annotated boundary are treated as aligned (Appendix~\ref{app:boundary-tolerance}). Thinking increases early submission to 94.1--99.7\% across seven models under the strict boundary, showing that additional reasoning can reinforce premature diagnosis submission.

\evmtakeaway{Models exhibit EVM: diagnosis submission timing is misaligned with evidence sufficiency. Greater capability does not eliminate it, and reasoning mode worsens it.}

\paragraph{RQ2: Is EVM caused by missing diagnostic knowledge?}\label{rq:belief}
To distinguish whether EVM reflects missing knowledge or a failure to act on evidence, we probe the ten candidate diagnoses of each \OURS{} case in cloned contexts without changing its original trajectory. Gold belief $p_{\text{gold}}$ is the probability of the ground-truth diagnosis normalized over all candidates. We measure $p_{\text{gold}}$ at diagnosis submission and at $t^*$. Since submitting a diagnosis terminates the diagnostic trajectory, boundary statistics include \textit{only trajectories remaining active at $t^*$}. We also conduct a thinking-mode sensitivity analysis for seven models and report expected calibration error (ECE). We group $N$ predictions into $M$ equal-width confidence bins $B_m$ using the highest candidate probability. Then $\mathrm{ECE}=\sum_{m=1}^{M}\frac{|B_m|}{N}|\operatorname{acc}(B_m)-\operatorname{conf}(B_m)|$, where $\operatorname{acc}$ and $\operatorname{conf}$ are the accuracy and mean confidence within each bin. Lower ECE indicates more reliable confidence (Appendix~\ref{app:mechanism-details}).

\begin{figure}[h!]
  \centering
  \begin{minipage}[t]{0.47\linewidth}
    \vspace{0pt}\centering
    \includegraphics[width=\linewidth,trim=0 2.5pt 0 0,clip]{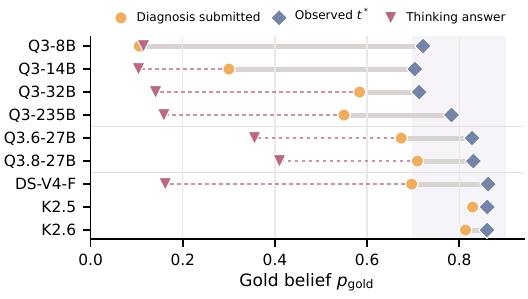}\vspace{-0.5em}
    \captionof{figure}{Gold belief from direct probes at diagnosis submission (circles) and $t^*$ (diamonds). Triangles use probes with thinking.}
    \label{fig:belief-commitment}
  \end{minipage}\hfill
  \begin{minipage}[t]{0.5\linewidth}
    \vspace{0pt}\centering
    \vspace{-0.7em}
    \captionof{table}{Effects of thinking on early submission, accuracy, gold belief, and ECE (\%). Accuracy uses answered cases. Off is gray. On is red for performance gains and blue otherwise.}\vspace{0.4em}
    \renewcommand{\arraystretch}{1.2}
    \setlength{\tabcolsep}{2mm}
    \resizebox{\linewidth}{!}{\definecolor{thinkingoff}{HTML}{65707B}
\definecolor{thinkingimproved}{HTML}{9B4350}
\definecolor{thinkingother}{HTML}{759AC3}
\begin{tabular}{lrrrrrrrr}
\toprule
\multirow{2}{*}{Model} & \multicolumn{2}{c}{Early $\downarrow$} & \multicolumn{2}{c}{Accuracy $\uparrow$} & \multicolumn{2}{c}{$p_{\mathrm{gold}}\uparrow$} & \multicolumn{2}{c}{ECE $\downarrow$} \\
\cmidrule(lr){2-3}\cmidrule(lr){4-5}\cmidrule(lr){6-7}\cmidrule(lr){8-9}
 & off & on & off & on & off & on & off & on \\
\midrule
Q3-8B & \textcolor{thinkingoff}{99.1} & \textcolor{thinkingother}{99.7} & \textcolor{thinkingoff}{10.9} & \textcolor{thinkingimproved}{11.3} & \textcolor{thinkingoff}{13.8} & \textcolor{thinkingother}{11.5} & \textcolor{thinkingoff}{65.4} & \textcolor{thinkingother}{89.8} \\
Q3-14B & \textcolor{thinkingoff}{78.6} & \textcolor{thinkingother}{99.7} & \textcolor{thinkingoff}{30.5} & \textcolor{thinkingother}{12.0} & \textcolor{thinkingoff}{28.7} & \textcolor{thinkingother}{10.4} & \textcolor{thinkingoff}{78.0} & \textcolor{thinkingother}{90.1} \\
Q3-32B & \textcolor{thinkingoff}{35.3} & \textcolor{thinkingother}{99.6} & \textcolor{thinkingoff}{59.2} & \textcolor{thinkingother}{13.9} & \textcolor{thinkingoff}{43.0} & \textcolor{thinkingother}{14.1} & \textcolor{thinkingoff}{55.0} & \textcolor{thinkingother}{87.9} \\
Q3-235B & \textcolor{thinkingoff}{49.7} & \textcolor{thinkingother}{99.6} & \textcolor{thinkingoff}{55.4} & \textcolor{thinkingother}{17.3} & \textcolor{thinkingoff}{52.6} & \textcolor{thinkingother}{15.9} & \textcolor{thinkingoff}{66.9} & \textcolor{thinkingother}{86.5} \\
Q3.6-27B & \textcolor{thinkingoff}{41.0} & \textcolor{thinkingother}{94.9} & \textcolor{thinkingoff}{70.2} & \textcolor{thinkingother}{37.8} & \textcolor{thinkingoff}{64.8} & \textcolor{thinkingother}{35.6} & \textcolor{thinkingoff}{36.3} & \textcolor{thinkingother}{76.1} \\
Q3.8-27B & \textcolor{thinkingoff}{36.6} & \textcolor{thinkingother}{94.1} & \textcolor{thinkingoff}{72.9} & \textcolor{thinkingother}{36.8} & \textcolor{thinkingoff}{64.2} & \textcolor{thinkingother}{41.0} & \textcolor{thinkingoff}{31.3} & \textcolor{thinkingother}{74.4} \\
DS-V4-F & \textcolor{thinkingoff}{24.6} & \textcolor{thinkingother}{97.3} & \textcolor{thinkingoff}{70.5} & \textcolor{thinkingother}{28.5} & \textcolor{thinkingoff}{43.9} & \textcolor{thinkingother}{16.2} & \textcolor{thinkingoff}{36.8} & \textcolor{thinkingother}{39.7} \\
\bottomrule
\end{tabular}
}
    \label{tab:thinking-sensitivity}
  \end{minipage}
\end{figure}

At observed $t^*$, mean gold belief $p_{\text{gold}}$ ranges from 0.70 to 0.86, showing that models can identify the correct diagnosis when sufficient evidence is available. $p_{\text{gold}}$ at diagnosis submission is more variable (Figure~\ref{fig:belief-commitment}). Thinking mode lowers submission-point gold belief and increases ECE in all seven models, indicating weaker support for the correct diagnosis and less reliable confidence when answers are submitted (Table~\ref{tab:thinking-sensitivity}). These findings suggest that models possess relevant diagnostic knowledge, but knowledge alone does not translate into appropriate diagnosis submission timing.

\evmtakeaway{EVM is not merely a knowledge deficit. Models can confidently identify the correct diagnosis from sufficient evidence yet submit at inappropriate times.}

\paragraph{RQ3: Does diagnosis submission depend on evidence order?}\label{rq:order}
For each case, we reorder only the clinical evidence presented before $t^*$ without changing any evidence content. The evidence at $t^*$ and all subsequent evidence retain their original order. Both sequences provide identical evidence sets at $t^*$. Diagnosis outcome disagreement occurs when the two trajectories submit different diagnoses, or one answers while the other remains unanswered. For pairs reaching $t^*$ in both orders, we compare probability distributions over the ten candidate diagnoses using Jensen--Shannon divergence (JSD). Lower JSD indicates more similar beliefs, with zero denoting identical distributions. For each metric, we average across permutations within each case and then across cases.
\begin{figure}[!ht]
  \centering
  \begin{minipage}[t]{0.42\linewidth}
    \vspace{0pt}\centering
    \includegraphics[width=\linewidth]{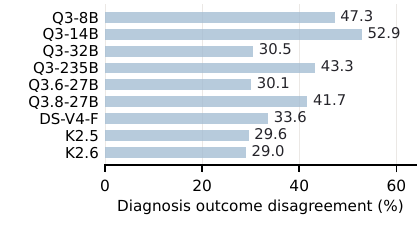}
  \end{minipage}\hspace{0.02\linewidth}
  \begin{minipage}[t]{0.54\linewidth}
    \vspace{0pt}
    \centering
    \includegraphics[width=\linewidth]{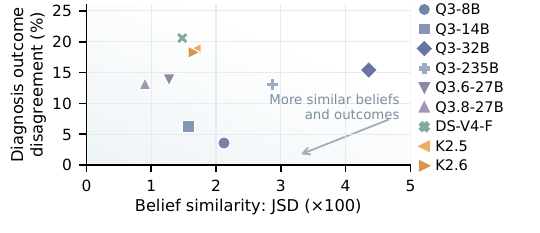}
  \end{minipage}
  \vspace{-1.2em}
  \caption{Diagnosis outcome disagreement after reordering in all cases (left), and its relation to belief similarity at $t^*$ among paired orders that both reach $t^*$ (right).}
  \label{fig:order-path}
\end{figure}

Reordering changes 29.0--52.9\% of final diagnosis outcomes across the full cohort (Figure~\ref{fig:order-path}, left). Among pairs reaching $t^*$ in both orders, mean JSD is low (0.009--0.044), yet substantial outcome disagreement remains (Figure~\ref{fig:order-path}, right). Thus, the evidence sequence affects final diagnoses even when both trajectories reach sufficient evidence and yield similar diagnostic beliefs (Appendix~\ref{app:mechanism-details}).

\evmtakeaway{Simply reordering the same evidence changes diagnosis outcomes despite similar diagnostic beliefs at the sufficiency boundary. Evidence order therefore affects how models act on available information.}

\paragraph{RQ4: How does misleading evidence change diagnosis submission?}\label{rq:strength}
We use two paired interventions to manipulate evidence value. First, we insert repeated clinical evidence or clinically irrelevant text into each case to test whether additions without new diagnostic information change diagnostic beliefs or decisions. Second, for 318 cases, we select an incorrect candidate diagnosis as the \textit{target diagnosis}. We construct weak and strong misleading evidence that differ in their support for this target while matching the evidence type, semantic direction, and format. We insert the paired versions at the same position early in the diagnostic trajectory, one turn before $t^*$, or one turn after $t^*$, with turns indexed in the augmented sequence. Target belief is the probability assigned to the incorrect target diagnosis, tracking confidence in this misleading alternative.

\begin{figure}[h!]
  \centering
  \begin{minipage}[t]{0.67\linewidth}
    \vspace{0pt}\centering
    \includegraphics[width=\linewidth,trim=0 2pt 0 0,clip]{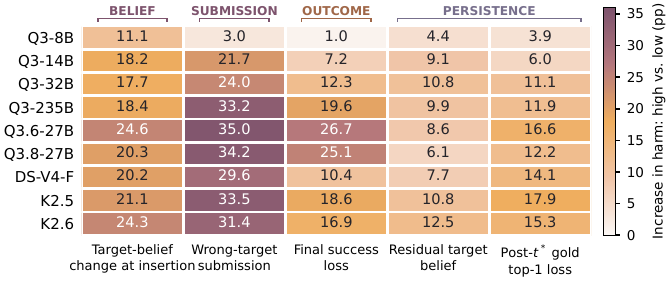}
    \vspace{-1.0mm}\scriptsize (a) Model-level propagation across endpoints.
  \end{minipage}\hfill
  \begin{minipage}[t]{0.31\linewidth}
    \vspace{0pt}\centering
    \includegraphics[width=\linewidth,trim=0 3pt 0 0,clip]{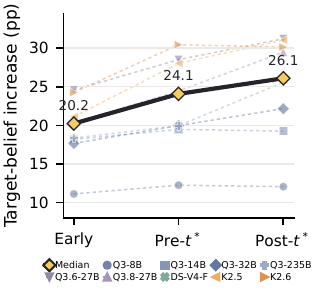}\vspace{0.1em}
    \scriptsize (b) Immediate response across stages.
  \end{minipage}\vspace{-0.3em}
  \vspace{-0.3em}
  \caption{Strong versus weak misleading evidence (pp, positive = harm). (a) From left: immediate target-belief gain, target submission rate increase, accuracy loss, target-belief gain after all evidence, and post-$t^*$ loss in correct diagnoses ranked first at completion. The first four use early insertion. (b) Immediate target-belief gains by insertion stage, with dashed curves for models and a solid median.}
  \label{fig:strength-cascade}
\end{figure}

The next diagnosis response differs in 5.5--34.4\% of comparisons between repeated-evidence and irrelevant-text branches, although neither adds new diagnostic information (Appendix~\ref{app:mechanism-details}). Stronger misleading evidence raises belief in the incorrect target and its diagnosis submission rate, reducing final diagnostic accuracy by up to 26.7 percentage points (Figure~\ref{fig:strength-cascade}a). The additional belief in the incorrect target persists after all original evidence arrives. Even after $t^*$, newly inserted misleading evidence shifts beliefs toward the incorrect target and reduces how often the correct diagnosis has the highest final probability.


\evmtakeaway{Models remain sensitive to misleading evidence even after evidence is sufficient. Stronger misleading evidence shifts beliefs toward incorrect diagnoses and causes persistent errors.}

\FloatBarrier
\endgroup

\section{Why EVM Matters: Predicting and Preventing Errors}
\label{sec:impact}
\begingroup
\setlength{\parskip}{4pt}
\setlength{\intextsep}{8pt plus 2pt minus 2pt}
\setlength{\textfloatsep}{10pt plus 2pt minus 2pt}

The value of EVM extends beyond describing model behavior. We test whether its features help predict submission timing and assess diagnostic errors, and whether reducing EVM improves accuracy.
We separate two prediction tasks: whether the next action submits a diagnosis after a \textsc{wait}, and whether a diagnosis is incorrect at its submission turn. The first uses checkpoints with an observed next turn, while the second uses each first submission. For each LLM, we fit regularized logistic regression with shared case-level splits and balance both classes within each split by majority-class undersampling. Baseline M1 uses (1) the fraction of evidence received, (2) evidence and candidate counts, and (3) diagnostic confidence, summarizing certainty across candidates. Our M3 adds (4) diagnosis submission propensity, capturing the tendency to submit rather than wait, and (5) belief instability, capturing changes in confidence between turns. AUROC~\citep{peterson1954theory} measures how well predictions distinguish submission from waiting, or incorrect from correct diagnoses, respectively. For next-turn submission, log loss (LL) evaluates predicted submission probabilities. A negative $\Delta\mathrm{LL}=\mathrm{LL}_{M3}-\mathrm{LL}_{M1}$ indicates better estimates of next-turn submission probabilities.

M3 improves AUROC for seven of nine LLMs in each task (Table~\ref{tab:impact-prediction}). Pooled AUROC rises from 0.835 to 0.854 for next-turn submission and from 0.845 to 0.879 for submission errors. EVM features thus inform both submission timing and error assessment beyond confidence alone. Next-turn log loss improves for eight models, with full baseline comparisons in Appendix~\ref{app:impact-details}.


\begin{figure}[!htb]
  \centering
  \begin{minipage}[t]{0.40\linewidth}
    \vspace{-0.6em}\centering
    \scriptsize
    \captionof{table}{AUROC for next-turn submission and errors at submission. M3 adds EVM features to M1. Higher is better.}\vspace{0.6em}
    \setlength{\tabcolsep}{2.1pt}
    \renewcommand{\arraystretch}{1.00}
    \begin{tabular*}{\linewidth}{@{\extracolsep{\fill}}lrrrr@{}}
\toprule
& \multicolumn{2}{c}{Next submission} & \multicolumn{2}{c}{Submission error} \\
\cmidrule(lr){2-3}\cmidrule(lr){4-5}
Model & M1 & M3 & M1 & M3 \\
\midrule
Q3-8B & 0.795 & 0.820 & 0.640 & 0.724 \\
Q3-14B & 0.780 & 0.827 & 0.627 & 0.822 \\
Q3-32B & 0.826 & 0.846 & 0.711 & 0.841 \\
Q3-235B & 0.827 & 0.840 & 0.789 & 0.847 \\
Q3.6-27B & 0.815 & 0.812 & 0.783 & 0.777 \\
Q3.8-27B & 0.857 & 0.885 & 0.773 & 0.789 \\
K2.5 & 0.851 & 0.842 & 0.800 & 0.800 \\
K2.6 & 0.863 & 0.873 & 0.711 & 0.723 \\
DS-V4-F & 0.790 & 0.807 & 0.814 & 0.839 \\
\midrule
Pooled & 0.835 & 0.854 & 0.845 & 0.879 \\
\bottomrule
\end{tabular*}

    \vspace{-0.0em}
    
    \label{tab:impact-prediction}\vspace{0.3em}
  \end{minipage}\hfill
  \begin{minipage}[t]{0.58\linewidth}
    \vspace{0.0em}\centering
    \includegraphics[width=\linewidth,trim=0 2pt 0 0,clip]{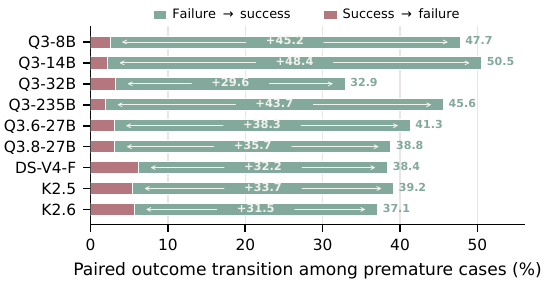}\vspace{-0.5em}
    \captionof{figure}{Natural Gate blocks pre-boundary diagnosis submission. Green shows failures rescued, red successes lost, and pale arrows net gains among early-answer cases.}
    \label{fig:impact-evidence}\vspace{0.3em}
  \end{minipage}
\end{figure}
To test whether reducing EVM changes diagnostic outcomes, we apply \textit{Natural Gate} to cases where the model originally answers before $t^*$. \textit{Natural Gate} replaces every attempted diagnosis submission before $t^*$ with \textsc{wait}, allowing further evidence to arrive. At $t^*$, we remove the gate and restore the original action prompt. The model then decides whether to wait or submit a diagnosis using the accumulated evidence. Among all intervened cases, 32.9--50.5\% change from failure to success, while far fewer change in the opposite direction (Figure~\ref{fig:impact-evidence}). Net diagnostic accuracy gains reach 29.6--48.4 percentage points across nine models, with every confidence interval above zero.

\evmtakeaway{EVM features help predict submission timing and assess diagnosis errors beyond confidence alone. Preventing premature submission can rescue failed diagnoses.}

\FloatBarrier
\endgroup

\section{\METHOD{}: Evidence-Verified Diagnosis Harness}
\label{sec:quorum}

\begin{figure}[t]
  \centering
  \vspace{-1.3em}\includegraphics[width=\linewidth]{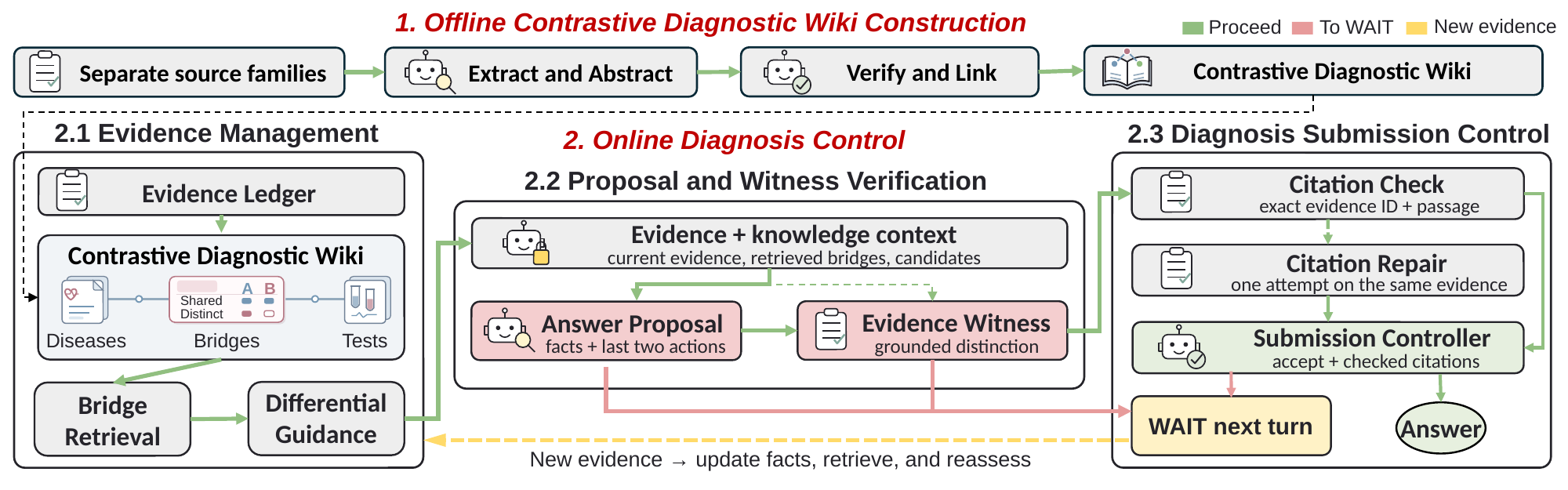}\vspace{-0.5em}
  \caption{\METHOD{} compiles independent case reports into contrastive knowledge, guides diagnosis, and authorizes answers grounded in current evidence.}
  \vspace{-1em}
  \label{fig:quorum-framework}
\end{figure}

\begingroup

Our EVM analyses reveal premature diagnosis submission and waiting after decisive evidence arrives. \METHOD{} decouples diagnosis generation from submission through an offline contrastive diagnostic Wiki and three online modules: evidence management, proposal and witness verification, and diagnosis submission control. Together, they manage the context and calls of a frozen LLM, using retrieved knowledge to guide diagnosis and current evidence to authorize its submission.

\subsection{Offline Diagnostic Wiki Construction}
\label{sec:evd-offline}
The \textit{Contrastive Diagnostic Wiki} (CDW) organizes clinical experience around findings that distinguish confusable diseases. We deduplicate published case reports using document identifiers and text similarity, excluding sources associated with \OURS{}, MedXpertQA, DiagnosisArena, and other existing benchmarks, including rewrites and derived questions. An LLM extracts the final diagnosis, competing diagnoses, shared presentations, and the findings used by the report authors to distinguish them. It retains the supporting passages and removes identifying details such as names and hospitals. Independent review checks each explanation against its supporting source passage. The resulting Wiki contains linked disease pages, test pages, and \textit{contrastive bridges}. Each bridge records confusable diseases, their shared presentation, a distinguishing finding, and its clinical context. Related entries are consolidated with source references into a Wiki fixed for evaluation.

\subsection{Online Diagnosis Control}
\label{sec:ecd-online}
\paragraph{Evidence Management}
For case $i$, turn $t$ begins when new clinical evidence $e_{it}$ arrives. The \textit{Evidence Ledger} appends it to the complete record $E_{i,\leq t}=(e_{i1},\ldots,e_{it})$. \textit{Bridge Retrieval} uses this record and the candidate set $\mathcal{Y}_i$ to retrieve disease and test pages from CDW, then follows their links to contrastive bridges. These pages form the knowledge view $K_{it}$, which explains the findings that distinguish competing diagnoses. \textit{Differential Guidance} provides $E_{i,\leq t}$ and $K_{it}$ to the diagnostic LLM as separately labeled patient evidence and reference knowledge. It asks the model to explain which competing diagnoses the current evidence distinguishes and which remain unresolved. This context supports the next diagnosis proposal or decision to wait.

\paragraph{Proposal and Witness Verification}
The \textit{Answer Proposal} calls the diagnostic LLM with $E_{i,\leq t}$, $K_{it}$, $\mathcal{Y}_i$, and an action history $H_{it}$ recording the harness actions from at most the two preceding turns. Keeping $E_{i,\leq t}$ complete allows reassessment using all available evidence, while limiting $H_{it}$ aims to prevent repeated waiting from dominating the next decision. The model returns $d_{it}\in\mathcal{Y}_i\cup\{\textsc{wait}\}$. A \textsc{wait} requests the next evidence unit. A proposed diagnosis instead triggers the \textit{Evidence Witness}, which reviews $d_{it}$ with the same evidence, knowledge, and candidates in a separate LLM context. It explains which findings support $d_{it}$ over competing diagnoses and identifies missing information that could change this choice. It returns an accept-or-wait verdict and citations $Q_{it}$ linking its explanation to patient evidence. Each citation $(j,q)$ pairs an evidence index $j$ with a passage $q$ copied from $e_{ij}\in E_{i,\leq t}$, so the cited text can be checked against the patient record.

\paragraph{Diagnosis Submission Control}
The \textit{Citation Check} reads each $(j,q)\in Q_{it}$, locates $e_{ij}$ in $E_{i,\leq t}$, and checks that $q$ is a nonempty passage copied exactly from that evidence unit. Let $\mathcal{P}_{it}$ denote all valid index--passage pairs in this record. If the \textit{Evidence Witness} accepts the diagnosis but citation checking fails, one \textit{Citation Repair} corrects the references using the same $E_{i,\leq t}$. The \textit{Submission Controller} then combines the checked citations with the witness verdict. Define $S_{it}=1$ when the witness accepts $d_{it}$, explains the distinguishing evidence, and identifies no remaining gap affecting the choice, and $S_{it}=0$ otherwise. The authorization status $V_{it}$ and first submission turn $\tau_i$ are
\begin{equation}
 V_{it}=S_{it}\,\mathbf{1}\!\left[\varnothing\neq Q_{it}\subseteq\mathcal{P}_{it}\right],
 \qquad \tau_i=\min\{t:d_{it}\in\mathcal{Y}_i,\ V_{it}=1\}.
\label{eq:evd-release}
\end{equation}
Here, $\mathbf{1}[\cdot]$ equals one when its condition holds and zero otherwise. Thus, $V_{it}=1$ requires both witness acceptance and a nonempty set of citations matching the observed patient record. At the first such turn, the harness submits $d_{i\tau_i}$ and ends the trajectory. Otherwise, it sends \textsc{wait}, appends this action to $H_{it}$, and repeats the modules when new evidence arrives. An initial \textsc{wait} sets $V_{it}=0$ directly. If no turn qualifies by evidence exhaustion, $\tau_i=\infty$ and the case is recorded as unanswered.

\subsection{Main Results}
\label{sec:evd-evaluation}
We compare the LLM-only baseline with \METHOD{} on \OURS{} using Qwen3.8-27B, DeepSeek-V4-Flash, Qwen3-32B, Kimi-K2.6, and Claude-Opus-4-8 \citep{anthropic2026opus48blog}. Unanswered cases count as errors for both methods. Following the RQ3 and RQ4 settings in Section~\ref{sec:mechanism}, we reorder evidence and insert duplicate, irrelevant, or paraphrased evidence into cases from \OURS{}, and vary misleading-evidence strength across 318 case pairs.

\begin{table}[!h]
\centering
\begingroup
\caption{LLM only versus \METHOD{} (\%). Non-answers count as errors. Premature answers precede $t^*$. Order flip counts diagnosis changes after reordering. Target error counts incorrect target submissions under strong misleading evidence before corrective evidence arrives.}
\label{tab:evc-outcomes-visual}
\definecolor{evcgray}{HTML}{A9B2BF}
\definecolor{evcmuted}{HTML}{757179}
\definecolor{evcblue}{HTML}{759AC3}
\definecolor{evcrow}{HTML}{F5F7F9}
\fontfamily{lmr}\fontsize{7.5}{9}\selectfont
\newcommand{\evccell}[2]{\includegraphics[width=53pt]{figures/evc_cell/evc_cell_#1_#2.pdf}}
\newcommand{\evcscale}[1]{\makebox[53pt]{\makebox[21pt]{#1}\makebox[32pt]{Value}}}
\textcolor{evcgray}{\rule{8pt}{4pt}} LLM only\qquad
\textcolor{evcblue}{\rule{8pt}{4pt}} LLM + \METHOD{}
\par\vspace{3pt}
{\fontfamily{lmr}\fontsize{7.5}{9}\selectfont
\begin{tblr}{width=\linewidth,colspec={@{}l X[c] X[c] X[c] X[c] X[c] X[c]@{}},colsep=1pt,rowsep=1pt,cells={valign=m},
  row{1}={font=\bfseries},row{2}={fg=evcmuted,font=\fontsize{6.5}{8}\selectfont},
  row{3,5,7}={bg=evcrow},hline{1,3,8}={0.4pt}}
Model & Accuracy $\uparrow$ & At $t^*$ $\uparrow$ & Premature $\downarrow$ & No answer $\downarrow$ & Order flip $\downarrow$ & Target error $\downarrow$ \\
 & \evcscale{0--100} & \evcscale{0--100} & \evcscale{0--60} & \evcscale{0--60} & \evcscale{0--50} & \evcscale{0--80} \\
Qwen3.8-27B & \evccell{0}{accuracy} & \evccell{0}{boundary} & \evccell{0}{premature} & \evccell{0}{noanswer} & \evccell{0}{order} & \evccell{0}{misleading} \\
DeepSeek-V4-Flash & \evccell{1}{accuracy} & \evccell{1}{boundary} & \evccell{1}{premature} & \evccell{1}{noanswer} & \evccell{1}{order} & \evccell{1}{misleading} \\
Qwen3-32B & \evccell{2}{accuracy} & \evccell{2}{boundary} & \evccell{2}{premature} & \evccell{2}{noanswer} & \evccell{2}{order} & \evccell{2}{misleading} \\
Kimi-K2.6 & \evccell{3}{accuracy} & \evccell{3}{boundary} & \evccell{3}{premature} & \evccell{3}{noanswer} & \evccell{3}{order} & \evccell{3}{misleading} \\
Claude-Opus-4-8 & \evccell{4}{accuracy} & \evccell{4}{boundary} & \evccell{4}{premature} & \evccell{4}{noanswer} & \evccell{4}{order} & \evccell{4}{misleading} \\
\end{tblr}
}
\endgroup
\end{table}

Accuracy improves for all five models, with fewer premature submissions and unanswered cases (Table~\ref{tab:evc-outcomes-visual}). The harness also reduces diagnosis changes after reordering and incorrect target submissions under strong misleading evidence. These gains combine improved diagnostic accuracy on the original cases with more consistent decisions under evidence perturbations.

\subsection{Additional Evaluations}

\paragraph{Component Ablations} On 200 cases from \OURS{}, we remove CDW knowledge or \textit{Evidence Witness}. Without the witness, proposed diagnoses are submitted without review. Removing it while retaining CDW lowers accuracy by 3.5--9.0 points. Removing CDW while retaining the witness lowers accuracy by 9.0--14.5 points (Table~\ref{tab:evd-ablation}). These ablations show that retrieved knowledge and evidence review each contribute to diagnostic success.

\begin{table}[!ht]
\centering
\definecolor{evdblue}{HTML}{3F6F9F}
\definecolor{evdgray}{HTML}{687482}
\begin{minipage}[t]{0.61\linewidth}
\vspace{-0.7em}
\caption{CDW and Evidence Witness ablations (\%). P: submission before $t^*$, N: no answer, A: accuracy.\strut}
\label{tab:evd-ablation}
\centering\fontsize{6.2}{8}\selectfont
\setlength{\tabcolsep}{0.8pt}
\begin{tabular*}{\linewidth}{@{\extracolsep{\fill}}l*{12}{r}@{}}
\toprule
\rule[-2pt]{0pt}{10pt}\multirow{2}{*}{Model} & \multicolumn{3}{c}{Full} & \multicolumn{3}{c}{$-$CDW} & \multicolumn{3}{c}{$-$Evidence Witness} & \multicolumn{3}{c}{$-$CDW,$-$Witness} \\
\cmidrule(lr){2-4}\cmidrule(lr){5-7}\cmidrule(lr){8-10}\cmidrule(lr){11-13}
\rule[-2pt]{0pt}{10pt}& P$\downarrow$ & N$\downarrow$ & A$\uparrow$ & P$\downarrow$ & N$\downarrow$ & A$\uparrow$ & P$\downarrow$ & N$\downarrow$ & A$\uparrow$ & P$\downarrow$ & N$\downarrow$ & A$\uparrow$ \\
\midrule
Q3.8-27B & \textcolor{evdblue}{15.0} & \textcolor{evdblue}{2.5} & \textcolor{evdblue}{75.5} & 47.0 & 6.5 & 63.0 & 19.5 & 0.5 & 72.0 & 52.5 & 2.5 & 57.5 \\
DS-V4-F & \textcolor{evdblue}{18.0} & \textcolor{evdblue}{1.5} & \textcolor{evdblue}{68.5} & 63.0 & 2.5 & 54.0 & 24.0 & 0.0 & 65.0 & 71.0 & 1.0 & 43.0 \\
Q3-32B & \textcolor{evdblue}{14.5} & \textcolor{evdblue}{6.0} & \textcolor{evdblue}{63.0} & 39.5 & 11.0 & 50.5 & 23.0 & 0.0 & 58.5 & 50.5 & 4.5 & 46.5 \\
Kimi-K2.6 & \textcolor{evdblue}{20.5} & \textcolor{evdblue}{3.0} & \textcolor{evdblue}{78.0} & 46.5 & 7.0 & 66.0 & 27.0 & 0.5 & 74.0 & 50.5 & 6.0 & 60.0 \\
Opus-4-8 & \textcolor{evdblue}{15.0} & \textcolor{evdblue}{8.5} & \textcolor{evdblue}{86.5} & 9.0 & 22.0 & 77.5 & 18.0 & 1.0 & 77.5 & 28.5 & 12.5 & 75.0 \\
\bottomrule
\end{tabular*}
\vspace{-0.7em}
\end{minipage}\hfill
\begin{minipage}[t]{0.37\linewidth}
\vspace{-0.7em}
\caption{Accuracy (\%) with alternative retrieval and full CDW.\strut}
\label{tab:evd-retrieval}
\centering\fontsize{6.2}{8}\selectfont
\setlength{\tabcolsep}{0.5pt}
\begin{tabular*}{\linewidth}{@{\extracolsep{\fill}}lrrrrr@{}}
\toprule
\rule[-2pt]{0pt}{10pt}\multirow{2}{*}{Model} & \multicolumn{5}{c}{Knowledge supplied to the harness} \\
\cmidrule(lr){2-6}
\rule[-2pt]{0pt}{10pt}& None & Random & BM25 & Contr. & CDW \\
\midrule
Q3.8-27B & 63.0 & 64.0 & 67.0 & 72.0 & \textcolor{evdblue}{75.5} \\
DS-V4-F & 54.0 & 55.0 & 59.0 & 65.0 & \textcolor{evdblue}{68.5} \\
Q3-32B & 50.5 & 51.5 & 54.5 & 59.5 & \textcolor{evdblue}{63.0} \\
Kimi-K2.6 & 66.0 & 67.0 & 70.0 & 75.0 & \textcolor{evdblue}{78.0} \\
Opus-4-8 & 77.5 & 78.5 & 80.5 & 84.5 & \textcolor{evdblue}{86.5} \\
\bottomrule
\end{tabular*}
\vspace{-0.7em}
\end{minipage}
\end{table}

\paragraph{Retrieval Controls} We hold other EVD-Harness modules fixed and vary only Wiki retrieval. \mbox{\textit{Random}} samples pages, while \textit{BM25} ranks them by lexical relevance to case evidence and candidates \mbox{\citep{robertson2009probabilistic}}. \textit{Contrastive} uses the same BM25 pages to show which findings are shared and which distinguish each candidate pair. \textit{Full CDW} follows linked pages over time and has the highest accuracy for all five models (Table~\ref{tab:evd-retrieval}). Further ablations appear in Appendix~\ref{app:evd-ablations}.

\FloatBarrier
\endgroup

\section{Conclusion}
\label{sec:conclusion}
We introduced \OURS{}, a progressive-evidence benchmark of 1,050 cases that evaluates diagnosis submission against annotated evidence sufficiency. Across nine LLMs, we identify EVM as a misalignment between diagnosis submission timing and evidence sufficiency. Even when models recognize the correct diagnosis, evidence order and misleading evidence redirect submission decisions. Scaling and thinking do not consistently resolve this mismatch. We then validate its diagnostic relevance through prediction and intervention: EVM signals impending errors, and preventing premature submission improves accuracy. To address this mismatch, we develop \METHOD{}, which retrieves contrastive diagnostic knowledge to guide diagnosis proposals, then authorizes submission through independent evidence review and citation checks. Across five LLMs, it improves accuracy by 12.0--51.1 percentage points and reduces sensitivity to evidence perturbations.

\section*{Limitations}
\label{sec:limitations}
First, \OURS{} is built from published case reports, so its disease distribution does not represent routine clinical prevalence. Broader sampling across care settings would strengthen external validity. Second, fixed candidates and scheduled evidence enable controlled comparisons, but active questioning and open-ended diagnosis remain to be studied. Third, human review covers only a subset of the sufficiency boundary annotations $t^*$. Model agreement and sampled audits support annotation quality, although individual boundaries may remain imprecise and warrant broader expert review.

\subsection*{AI use statement}

We used AI in three parts of this work: (1) to polish the wording of the manuscript, with all initial drafts written by the authors; (2) to assist with data synthesis and filtering during benchmark construction, followed by human quality control, as detailed in Appendix~\ref{app:benchmark-construction}; and (3) as the models evaluated in our experiments. We reviewed all AI-assisted work and take responsibility for the final content of this paper and its associated artifacts.

\subsection*{Ethics statement}

Our study of evidence--value misalignment in language models does not raise specific ethical concerns beyond data privacy. Although the benchmark draws on real clinical case reports, patient-identifying information had already been removed when these reports were published. We also checked for potentially identifying information during data screening to verify that the selected cases preserved patient privacy.

\subsection*{Reproducibility statement}

The appendices describe benchmark construction and all experimental settings in detail to support reproducibility. We will release the code for all experiments, including the implementation of \METHOD{}, to enable others to reproduce our results.



\bibliography{iclr2027_conference}
\bibliographystyle{iclr2027_conference}

\appendix
\setcounter{table}{0}
\setcounter{figure}{0}
\setcounter{section}{0}
\setcounter{equation}{0}
\newcounter{promptcounter}
\setcounter{promptcounter}{0}
\renewcommand{\thetable}{A\arabic{table}}
\renewcommand{\thefigure}{A\arabic{figure}}
\renewcommand{\thesection}{A\arabic{section}}
\renewcommand{\theequation}{A\arabic{equation}}
\renewcommand{\theHtable}{appendix.\arabic{table}}
\renewcommand{\theHfigure}{appendix.\arabic{figure}}
\renewcommand{\theHequation}{appendix.\arabic{equation}}
\renewcommand{\theHsection}{appendix.\arabic{section}}
\section{Detailed Construction of the \OURS{} Benchmark}
\label{app:benchmark-construction}

This appendix details the data sources, evidence and candidate construction, sufficiency annotation, and quality validation underlying \OURS{}, as summarized in Section~\ref{sec:benchmark}. We complete all item filtering and annotation before evaluating the target models and make no subsequent revisions based on the reported evaluation results.

\subsection{Case Collection and Structural Filtering}

We begin with a pool of 75,434 English-language medical case reports assembled from multiple sources. These include case collections used in existing medical benchmarks, such as DiagnosisArena \citep{zhu2026diagnosisarena}, and journal reports acquired or crawled from PubMed-indexed sources \citep{white2020pubmed}. The source journals include NEJM, Cureus, BMJ Case Reports, Gastroenterology, JAMA, Neurology, The Lancet, Medicine, and Chest, among others. Every case in the final benchmark retains a case ID linking it to the original report pool and a journal-source label identifying its publication venue.

Structural filtering proceeds in three steps. First, we normalize each narrative into four fields: case information, physical examination, diagnostic tests, and final diagnosis. This step yields 52,147 structured records. Second, we check basic completeness, requiring at least 300 characters of clinical history, a chief complaint or symptoms, non-empty diagnostic-test text, and a single normalized final diagnosis. Third, we assess whether the report contains sufficient evidence to support a progressive diagnostic sequence. Each report must include at least two laboratory, imaging, pathology, or special-test findings, with evidence available at all three stages of presentation, basic workup, and advanced confirmation. Across these checks, we exclude 6,378 reports with insufficient history, 8,130 with incomplete evidence stages, 1,437 with fewer than two tests, 94 without diagnostic-test text, and 88 without a chief complaint or symptoms. The remaining 36,020 reports proceed to evidence and option construction.

\subsection{Evidence and Option Construction}

For each qualified report, we retain information available before diagnosis and remove explicit diagnosis names, treatment outcomes, and follow-up information. We then segment the remaining narrative into minimal, self-contained units of clinical evidence. Each evidence item retains an anchor to the original text, its source section, its clinical stage, and an evidence-type label. These labels cover demographics, chief complaint, history, symptoms, medication, vital signs, physical examination, laboratory tests, imaging, pathology, procedures, and special tests. We arrange the evidence items in the order of clinical disclosure in the source report. This preserves a traceable progression of available information and prevents model-generated reordering intended solely to increase task difficulty.

We convert each case into a ten-way diagnostic question comprising the confirmed diagnosis and nine clinically plausible confounders. GPT-5.4 first proposes 10--15 differential diagnoses, from which it selects nine based on clinical similarity and whether the complete case evidence can rule them out. For each confounder, the model must identify at least three shared symptoms or signs that make it plausible early in the case, specific evidence from the report that supports it, and a test or finding that ultimately excludes it. We discard candidates that are synonymous with the gold diagnosis, differ only in diagnostic granularity, lack support in the case, or cannot be ruled out by the available evidence. We remove explicit diagnostic wording from the question stem and randomize the position of the correct answer.

We next screen the constructed questions for answer uniqueness and difficulty. The complete evidence must support exactly one candidate, so we exclude items with compound diagnoses, multiple defensible answers, diagnostic leakage, or inconsistent option granularity. Multiple strong models then solve each question independently. Items that can be answered consistently from superficial cues without integrating the clinical evidence are excluded as too easy. This stage retains 7,472 usable multiple-choice questions for sufficiency-boundary annotation and quality validation.

\subsection{Sufficiency Boundary and Quality Validation}

The boundary-annotation prompt contains the ten diagnostic candidates and the ordered sequence of atomic evidence. Claude Opus 4.8, Gemini 3.1 Pro Preview, and GPT-5.6-sol independently identify the earliest position at which the accumulated evidence uniquely distinguishes the correct diagnosis from all competing candidates. We retain only cases for which all three annotators identify exactly the same position and assign this position as $t_i^*$. This procedure operationalizes the sufficiency criterion in Section~\ref{sec:benchmark}, providing a case-level reference independent of the evaluated model.

Automated quality control then reviews the source case, candidate set, and proposed boundary in a fixed sequence. We first recheck the source report for sufficient clinical detail, internal consistency, and a definitive diagnosis, removing cases with unresolved deficiencies. We next verify that the complete evidence supports a unique answer and exclude any remaining low-difficulty items that do not require evidence integration. Finally, Gemini 3.5 Flash independently validates the proposed boundary against the candidates and ordered evidence sequence. We obtain three independent judgments per item and retain it only if all three confirm that $t_i^*$ is the earliest sufficient position. Gemini 3.5 Flash serves as a separate validator and does not contribute to the preceding three-model boundary agreement.

To assess the quality of the automated annotations, five graduate researchers with medical backgrounds independently review 200 randomly sampled items. They evaluate the coherence of each case, the uniqueness of its correct answer, the clinical plausibility of its confounders, and whether $t_i^*$ marks the earliest sufficient boundary. Item-level inter-rater agreement reaches 92.5\%, and majority voting rates 94.0\% of the sampled items as accurately annotated. Across cases, the mean within-item standard deviation of the human boundary annotations is 0.84 turns, showing modest dispersion in the judged position across reviewers. We examine sensitivity to one- and two-turn tolerance windows in Appendix~\ref{app:boundary-tolerance}.

Quality control yields 1,810 valid dynamic-diagnosis items. From this pool, we select 1,050 cases through stratified sampling by disease system and empirical difficulty to form the final \OURS{} benchmark. The release covers 953 distinct diagnoses across 24 disease systems and contains 26,869 atomic evidence items. Each case includes ten diagnostic candidates and an average of 25.6 evidence items, and the mean position of the sufficiency boundary $t_i^*$ is 21.3.

\section{Metric Definitions}
\label{app:metric-definitions}

This appendix defines the metrics used to characterize EVM, compare diagnostic beliefs, and evaluate interventions. Each group is linked to the experiment it supports. For case $i$, let $\mathcal{Y}_i$ be the ten candidate diagnoses, $y_i^*$ the correct diagnosis, $T_i$ the number of evidence items, and $t_i^*$ the annotated sufficiency boundary. A turn is indexed by the number of evidence items received. Probabilities and rates lie in $[0,1]$ and are multiplied by 100 when reported as percentages. Differences between percentages are reported in percentage points (pp).

\subsection{Diagnosis Outcomes and Submission Timing}
\label{app:metric-outcomes}

\paragraph{Accuracy and submission rate.}
The timing analysis in \hyperref[rq:timing]{RQ1} and the intervention analyses in Section~\ref{sec:impact} distinguish whether a model submits a diagnosis from whether that diagnosis is correct. Let $\tau_i$ be its first submission turn, with $\tau_i=\infty$ if it reaches the end of the evidence without submitting. Let $A_i=\mathbf{1}[\tau_i<\infty]$, and let $Z_i=1$ for a correct submitted diagnosis and zero otherwise. For an evaluated set of $N$ cases,
\begin{equation}
\mathrm{Acc}_{\mathrm{submission}}=\frac{\sum_i Z_i}{\sum_i A_i},
\qquad
\mathrm{Acc}_{\mathrm{all}}=\frac{1}{N}\sum_i Z_i,
\qquad
\mathrm{Submission\ rate}=\frac{1}{N}\sum_i A_i.
\label{eq:accuracy-definitions}
\end{equation}
Submission accuracy evaluates only cases with a submitted diagnosis. Accuracy over all cases, also called diagnostic success, counts both incorrect submissions and non-answers as errors. The thinking comparison in \hyperref[rq:belief]{RQ2} reports submission accuracy. The misleading-evidence, Natural Gate, and harness evaluations use accuracy over all evaluated cases. The submission rate, also called answer coverage, shows how often a model produces any diagnosis. If no diagnosis is submitted, submission accuracy is undefined.

\paragraph{Timing relative to sufficient evidence.}
\hyperref[rq:timing]{RQ1} assigns each trajectory to one of four categories:
\begin{equation}
\begin{aligned}
R_{\mathrm{pre}}&=\frac{1}{N}\sum_i\mathbf{1}[\tau_i<t_i^*], &
R_{\mathrm{at}}&=\frac{1}{N}\sum_i\mathbf{1}[\tau_i=t_i^*],\\
R_{\mathrm{late}}&=\frac{1}{N}\sum_i\mathbf{1}[t_i^*<\tau_i<\infty], &
R_{\mathrm{none}}&=\frac{1}{N}\sum_i\mathbf{1}[\tau_i=\infty].
\end{aligned}
\label{eq:timing-rates}
\end{equation}
These are the rates of premature submission, submission at the boundary, delayed submission, and no submission. They use all cases as the denominator and sum to one. Timing and correctness are separate properties: a correct diagnosis can be premature, and a diagnosis submitted after the boundary can be incorrect. Keeping non-answers separate also distinguishes eventual submission after sufficient evidence from waiting through the complete record.

\subsection{Diagnostic Beliefs, Action Probabilities, and Calibration}
\label{app:metric-beliefs}

\paragraph{Candidate and action probabilities.}
\hyperref[rq:belief]{RQ2} measures the distribution $p_{it}$ over candidate diagnoses at evidence checkpoint $t$. For the default JSON-prefix measurement in either thinking mode, let $\ell_{it}(y)$ be the conditional log-probability of the candidate label in the diagnosis field, and let $u_{it}(a)$ be that of action $a$ in the action field. Each score sums the conditional log-probabilities of the tokens in the corresponding label or action. The normalized probabilities are
\begin{equation}
 p_{it}(y)=\frac{\exp\ell_{it}(y)}{\sum_{y'\in\mathcal{Y}_i}\exp\ell_{it}(y')},
 \qquad
 p_{it}(\mathrm{answer})=\frac{\exp u_{it}(\mathrm{answer})}
 {\exp u_{it}(\mathrm{answer})+\exp u_{it}(\mathrm{wait})}.
\label{eq:appendix-candidate-probability}
\end{equation}
The first distribution measures which diagnosis the model favors at the fixed answer-field prefix. The second measures its propensity to submit a diagnosis at the action-field prefix. In all conditions, scoring uses a cloned context containing the candidate list, available evidence, and preceding conversation. The candidate-scoring prefix is the same whether the original action is \textsc{answer} or \textsc{wait}, and probe outputs are not added to the ongoing trajectory. With thinking enabled, probabilities also condition on the generated reasoning. Appendix~\ref{app:mechanism-protocol} specifies the prefixes, and Appendix~\ref{app:thinking-measurement} illustrates their use in both modes.

\paragraph{Correct and target diagnoses.}
Gold belief is the probability assigned to the correct diagnosis. In \hyperref[rq:strength]{RQ4}, target belief is the probability assigned to the prespecified incorrect diagnosis supported by the inserted evidence:
\begin{equation}
 p_{\mathrm{gold},it}=p_{it}(y_i^*),
 \qquad p_{\mathrm{target},it}=p_{it}(y_i^{\mathrm{tar}}).
\label{eq:gold-belief}
\end{equation}
The difference $p_{\mathrm{gold}}-p_{\mathrm{target}}$ is positive when the correct diagnosis is favored over the target. A correct top-1 prediction means that the highest-probability candidate is the correct diagnosis. Its frequency measures diagnostic ranking at a specified evidence prefix, which need not equal the accuracy of diagnoses submitted in a natural trajectory. RQ2 reports mean gold belief at submission over answered cases and at $t_i^*$ over cases that reach that turn. These two means can therefore involve different sets of cases.

\paragraph{Confidence and calibration.}
Let $p_{it}^{(1)}\geq p_{it}^{(2)}$ be the two largest candidate probabilities. Top-1 confidence is $p_{it}^{(1)}$, and the probability margin is $p_{it}^{(1)}-p_{it}^{(2)}$. A large margin indicates a clear preference for the leading candidate, but does not establish that it is correct. The thinking comparison in \hyperref[rq:belief]{RQ2} evaluates calibration using ten equal-width bins over $[0,1]$. For $n$ measured predictions, with bin $B_m$ containing predictions whose top-1 confidence falls in that interval,
\begin{equation}
\mathrm{ECE}=\sum_{m=1}^{10}\frac{|B_m|}{n}
\left|\operatorname{acc}(B_m)-\operatorname{conf}(B_m)\right|.
\label{eq:ece}
\end{equation}
Here, $\operatorname{acc}(B_m)$ is the fraction of correct top-1 predictions and $\operatorname{conf}(B_m)$ their mean top-1 confidence. Empty bins contribute zero, and the last bin includes confidence one. Lower expected calibration error (ECE) means closer agreement between confidence and observed correctness. The thinking table pools all measured evidence checkpoints for ECE and margin, while its gold-belief column uses submission turns.

\subsection{Evidence-Order and Misleading-Evidence Effects}
\label{app:metric-interventions}

\paragraph{Distribution and diagnosis disagreement (\hyperref[rq:order]{RQ3}).}
For candidate distributions $P$ and $Q$ from the two evidence orders, let $M=(P+Q)/2$. Jensen--Shannon divergence is
\begin{equation}
\operatorname{JSD}(P,Q)=\frac12\sum_yP(y)\log\frac{P(y)}{M(y)}
 +\frac12\sum_yQ(y)\log\frac{Q(y)}{M(y)}.
\label{eq:jsd}
\end{equation}
We use natural logarithms and take zero-probability terms to contribute zero. JSD ranges from zero to $\log2$, with zero indicating identical distributions. RQ3 measures it at $t_i^*$ only for pairs in which both orders reach that turn, so both distributions are based on the same evidence set.

Diagnosis outcome disagreement is the indicator that two trajectories submit different diagnoses or that only one submits a diagnosis. Two non-answers count as agreement. RQ3 reports its mean both over all evaluated pairs and over the pairs used for boundary JSD. For each metric, eligible permutations are first averaged within a source case and then across cases. This gives each source case equal weight. The candidate-label control uses the same outcome comparison after mapping labels back to diagnoses. The repeated-evidence control instead compares the next response after insertion at a shared evidence prefix, so its disagreement rate describes a local response rather than the outcome of a complete trajectory.

\paragraph{Effects of stronger misleading evidence (\hyperref[rq:strength]{RQ4}).}
Each comparison fixes the source case, incorrect target diagnosis, and insertion position. Write S and W for the strong and weak versions. The additional change in target belief immediately after insertion is
\begin{equation}
\Delta p_{\mathrm{target}}=
\bigl(p_{\mathrm{target,after}}^{\mathrm{S}}-p_{\mathrm{target,before}}^{\mathrm{S}}\bigr)
-\bigl(p_{\mathrm{target,after}}^{\mathrm{W}}-p_{\mathrm{target,before}}^{\mathrm{W}}\bigr).
\label{eq:target-belief-change}
\end{equation}
The remaining belief effect is $p_{\mathrm{target,final}}^{\mathrm{S}}-p_{\mathrm{target,final}}^{\mathrm{W}}$, measured after all evidence has arrived. Target submission rate is the fraction of evaluated trajectories that submit the incorrect target, and its increase is the strong-minus-weak difference. Diagnostic accuracy loss is $\mathrm{Acc}_{\mathrm{all}}^{\mathrm{W}}-\mathrm{Acc}_{\mathrm{all}}^{\mathrm{S}}$. Correct-top-1 loss is the weak-minus-strong difference in the fraction ranking the correct diagnosis first at sequence completion. All five effects are oriented so that positive values indicate harm from stronger misleading evidence. Figure~\ref{fig:strength-cascade}a uses early insertion for the first four effects and insertion after the original boundary for correct-top-1 loss.

\subsection{Error Prediction and Natural Gate Outcomes}
\label{app:metric-impact}

\paragraph{Submission prediction and error assessment (Section~\ref{sec:impact}).}
For next-turn submission prediction, an eligible checkpoint is a wait followed by an observed next turn. Its label $r_j$ is one if the next action submits any diagnosis and zero if it continues waiting. For error assessment at submission, examples are first submissions, with $r_j=1$ for an incorrect diagnosis and zero for a correct diagnosis. The regression output $q_j$ estimates the probability of the positive outcome for the corresponding task. AUROC measures how often a positive example receives a higher score than a negative example, counting ties as one half. A value of 0.5 indicates chance ranking and one indicates perfect ranking. Log loss evaluates the probabilities:
\begin{equation}
\mathrm{LL}=-\frac{1}{n}\sum_{j=1}^{n}
\left[r_j\log q_j+(1-r_j)\log(1-q_j)\right].
\label{eq:log-loss}
\end{equation}
Lower log loss is better. For next-turn submission, we report $\Delta\mathrm{LL}=\mathrm{LL}_{\mathrm{M3}}-\mathrm{LL}_{\mathrm{M1}}$, so negative values favor M3. Metrics use class-balanced held-out examples. Appendix~\ref{app:impact-prediction-protocol} describes the case-level splits, majority-class undersampling, and regression fitting.

\paragraph{Errors corrected and correct diagnoses lost (Section~\ref{sec:impact}).}
Natural Gate pairs the original and gated outcomes of every case with an original premature submission. Let $N_{\mathrm{pre}}$ be the number of such cases and $Z_i^{(0)},Z_i^{(1)}$ their original and gated success indicators. Then
\begin{equation}
\begin{aligned}
R_{0\rightarrow1}&=\frac{1}{N_{\mathrm{pre}}}\sum_i\mathbf{1}[Z_i^{(0)}=0,Z_i^{(1)}=1],\\
R_{1\rightarrow0}&=\frac{1}{N_{\mathrm{pre}}}\sum_i\mathbf{1}[Z_i^{(0)}=1,Z_i^{(1)}=0],\\
\Delta\mathrm{Acc}&=R_{0\rightarrow1}-R_{1\rightarrow0}
=\mathrm{Acc}_{\mathrm{all}}^{(1)}-\mathrm{Acc}_{\mathrm{all}}^{(0)}.
\end{aligned}
\label{eq:paired-transitions}
\end{equation}
The first two rates are the percentages of errors corrected and correct diagnoses lost in Table~\ref{tab:impact-gate-full}. Both use all selected cases as the denominator. Their difference is the paired accuracy gain, including failures to submit under the gate. The gain over the complete benchmark is $(N_{\mathrm{pre}}/1050)\Delta\mathrm{Acc}$, with nonselected cases retaining their original outcomes. Appendix~\ref{app:impact-natural-gate} describes the intervention and gives worked examples.

\paragraph{Confidence intervals.}
Reported paired intervals use 2,000 bootstrap resamples of source cases. All checkpoints, targets, permutations, or intervention versions belonging to a sampled case remain together, and each contrast is recomputed within the resample. The 2.5th and 97.5th percentiles give the 95\% interval. For prediction, this resampling uses held-out cases and compares M1 and M3 on identical checkpoints. For interventions, it preserves the original--intervention pairing.

\section{Experimental Details and Additional Mechanism Analyses}
\label{app:mechanism-details}

\definecolor{appBhead}{RGB}{255,255,255}
\definecolor{appBstripe}{RGB}{255,255,255}
\definecolor{appBink}{RGB}{50,78,112}
\definecolor{appBmuted}{RGB}{104,116,131}

\subsection{Shared Interaction and Probability Measurement}
\label{app:mechanism-protocol}

The four mechanism experiments in Section~\ref{sec:mechanism} share a sequential diagnosis protocol. The model receives ten candidate diagnoses, followed by one evidence item per turn. Each request includes the initial instructions, candidates, accumulated evidence, and previous JSON actions. It returns \verb|{"action":"wait","answer":""}| to receive more evidence or \verb|{"action":"answer","answer":"<option letter>"}| to submit a diagnosis. The first submission ends the trajectory. Waiting after the final evidence item produces a non-answer. Accuracy, submission rate, and timing categories follow Appendix~\ref{app:metric-outcomes}.

Figure~\ref{fig:diagnosis-interaction-prompt} reproduces the system prompt used with thinking disabled and the templates for the case-opening and evidence messages. The model is instructed to answer once the available evidence supports a reliable choice. Neither the ground-truth diagnosis nor $t_i^*$ is included in its input.

\begin{figure}[!ht]
\centering
\begingroup
\definecolor{mechanismboxbg}{RGB}{247,249,252}
\definecolor{mechanismboxline}{RGB}{156,173,198}
\setlength{\fboxsep}{9pt}
\setlength{\fboxrule}{0.6pt}
\fcolorbox{mechanismboxline}{mechanismboxbg}{%
\begin{minipage}{\dimexpr\linewidth-2\fboxsep-2\fboxrule\relax}
\footnotesize\raggedright
\textbf{SYSTEM PROMPT}\par\smallskip
You are the doctor agent in a closed-set, multi-turn diagnosis task.\par
A patient agent will reveal exactly one new piece of clinical evidence per turn. After each piece, choose one action and output JSON only:\par\smallskip
\texttt{\{"action":"wait","answer":""\}}\quad or\par
\texttt{\{"action":"answer","answer":"<option letter>"\}}\par\smallskip
Decision rule:\par
1. Choose "wait" only when the evidence available so far does not uniquely identify one diagnosis among the listed candidates.\par
2. Choose "answer" as soon as the accumulated evidence rules out all other listed candidates and uniquely identifies one diagnosis. You do not need to wait for every piece of evidence.\par
3. This is a closed-set candidate-selection task, not a real clinical workup. Once alternatives can be excluded with adequate confidence, answer.\par
4. An answer is final. Its option letter must be one of the listed candidates. When waiting, answer must be empty.\par
5. Use only evidence revealed in this conversation.\par
6. Do not provide confidence, explanation, markdown, extra keys, or chain-of-thought.\par
\medskip\textcolor{mechanismboxline}{\rule{\linewidth}{0.4pt}}\par\medskip
\textbf{CASE-OPENING MESSAGE}\par\smallskip
A patient case needs diagnosis. There are 10 candidate diagnoses:\par
\texttt{A. [diagnosis A]}\quad\ldots\quad\texttt{J. [diagnosis J]}\par
The patient will now provide one new atomic piece of evidence at a time.\par\medskip
\textbf{EVIDENCE MESSAGES}\par\smallskip
First turn: \texttt{Patient evidence [evidence ID]: [evidence text]}\par
Later turns: \texttt{New patient evidence [evidence ID]: [evidence text]}
\end{minipage}}
\endgroup
\caption{Diagnosis interaction prompt. The system prompt is reproduced verbatim, with line wrapping adjusted for readability. Bracketed fields in the message templates are filled from the case record, and the complete candidate list is provided at the start.}
\label{fig:diagnosis-interaction-prompt}
\end{figure}

The paired interventions retain the source case, candidate diagnosis texts, correct diagnosis, and generation settings. RQ3 changes evidence order, while RQ4 inserts repeated, irrelevant, or misleading evidence. A supplementary control changes candidate labels. Each design below specifies whether it evaluates a natural trajectory or a reconstructed evidence prefix.

\paragraph{Conversation state for measurement.}
At each measured turn, we copy the current conversation containing the candidate list, all evidence received so far, and the preceding actions. In this cloned context, we fix the scoring prefix and evaluate candidate continuations. The same procedure applies at diagnosis-submission and waiting turns. Neither the scoring prefix nor its continuations are added to the ongoing conversation. The original action therefore determines whether the diagnostic trajectory ends or receives further evidence.

\paragraph{Candidate probabilities.}
Candidate scoring uses the fixed diagnosis prefix \verb|{"action":"answer","answer":"| in all conditions. At this position, we score all ten candidate labels as alternative continuations of the same context and normalize them using Equation~\ref{eq:appendix-candidate-probability}. For a multi-token label, its log-probability sums the conditional log-probabilities of its tokens. With thinking enabled, the scoring context also contains the reasoning generated before the JSON response.

\paragraph{Action probabilities.}
Action scoring uses the shorter prefix \verb|{"action":"|. We score the complete continuations \texttt{wait} and \texttt{answer} and normalize over the two actions as in Equation~\ref{eq:appendix-candidate-probability}. The resulting $p(\mathrm{answer})$ measures submission propensity, while the candidate distribution measures diagnostic belief over the ten diagnoses at the same evidence prefix.

Each model is evaluated on the same 1,050 cases. Paired confidence intervals use the case-bootstrap procedure in Appendix~\ref{app:metric-impact}. Natural measurements end at the first diagnosis submission, so later-turn summaries identify which trajectories remain available.

\subsection{Diagnosis Submission Timing and Belief at the Sufficiency Boundary}

In \hyperref[rq:timing]{RQ1}, each case follows its original evidence order. Figure~\ref{fig:evm-phenotypes} reports the four submission-timing categories defined in Appendix~\ref{app:metric-outcomes}, separately from diagnostic correctness.

In \hyperref[rq:belief]{RQ2}, we measure candidate probabilities at each evidence checkpoint and summarize $p_{\mathrm{gold}}$ at submission and at $t_i^*$. The submission summary includes answered trajectories. The boundary summary includes trajectories that reach $t_i^*$, whether they answer or wait at that turn. Figure~\ref{fig:belief-commitment} therefore compares belief at actual submission with belief upon receiving sufficient evidence, using the cases available at each point, which may differ because early submissions end before sufficient evidence arrives.

\paragraph{Sensitivity to boundary tolerance.}
\label{app:boundary-tolerance}
The human review reports a mean within-case standard deviation of 0.84 turns for boundary judgments (Appendix~\ref{app:benchmark-construction}). We therefore examine whether the RQ1 patterns depend on an exact turn match. Keeping each annotated $t_i^*$ as the reference, we rescore the existing trajectories with tolerance $k\in\{0,1,2\}$. A submission is premature if $\tau_i<t_i^*-k$, aligned if $|\tau_i-t_i^*|\leq k$, and delayed if $\tau_i>t_i^*+k$. Unanswered cases remain a separate category. This changes the timing score without changing the evidence, model responses, or diagnostic correctness. The window accommodates uncertainty in the annotation rather than asserting that evidence before $t_i^*$ is sufficient.

\begin{table}[htbp]
\centering\small
\caption{RQ1 timing sensitivity to a tolerance of $k$ turns around the annotated boundary. All rates use the same 1,050 cases per model. Non-answer rates are unchanged and omitted.}\vspace{0.5em}
\label{tab:boundary-tolerance}
\setlength{\tabcolsep}{4mm}
\resizebox{\linewidth}{!}{\begin{tabular}{lrrrrrrrrr}
\toprule
& \multicolumn{3}{c}{Premature (\%)} & \multicolumn{3}{c}{Within window (\%)} & \multicolumn{3}{c}{Delayed (\%)}\\
\cmidrule(lr){2-4}\cmidrule(lr){5-7}\cmidrule(lr){8-10}
Model & $k=0$ & $k=1$ & $k=2$ & $k=0$ & $k=1$ & $k=2$ & $k=0$ & $k=1$ & $k=2$\\
\midrule
Q3-8B & 99.1 & 97.7 & 96.0 & 0.6 & 2.2 & 3.9 & 0.3 & 0.1 & 0.1\\
Q3-14B & 78.6 & 70.3 & 63.7 & 17.6 & 27.1 & 34.2 & 2.7 & 1.4 & 1.0\\
Q3-32B & 35.3 & 27.0 & 21.6 & 37.0 & 48.5 & 55.7 & 9.6 & 6.5 & 4.7\\
Q3-235B & 49.7 & 40.2 & 33.5 & 38.3 & 49.9 & 58.6 & 6.8 & 4.7 & 2.7\\
Q3.6-27B & 41.0 & 34.8 & 28.7 & 42.7 & 51.9 & 59.0 & 6.4 & 3.4 & 2.5\\
Q3.8-27B & 36.6 & 26.8 & 21.0 & 45.6 & 59.0 & 66.7 & 8.9 & 5.2 & 3.4\\
DS-V4-F & 24.6 & 20.8 & 17.4 & 27.0 & 35.0 & 41.5 & 11.0 & 6.8 & 3.6\\
K2.5 & 15.8 & 11.6 & 9.4 & 43.7 & 50.7 & 54.3 & 6.8 & 4.0 & 2.6\\
K2.6 & 17.0 & 12.9 & 10.7 & 44.3 & 51.6 & 55.3 & 7.9 & 4.7 & 3.1\\
\bottomrule
\end{tabular}
}
\end{table}

Allowing one turn increases the aligned fraction by 1.6--13.4 percentage points across models, and allowing two turns increases it by 3.3--21.0 points (Table~\ref{tab:boundary-tolerance}). These increases reflect the more permissive scoring rule. They do not remove the main behavioral differences: Qwen3-8B still submits prematurely in 96.0\% of cases at $k=2$, while DeepSeek and the Kimi models retain non-answer rates of 30.9--37.4\%. The tolerance reduces both premature and delayed classifications, but substantial timing mismatch or non-response remains for every model. We retain $k=0$ for the other reported experiments so that their evidence checkpoints and intervention positions share the same annotated reference.

Applying the same tolerance to the thinking trajectories yields premature-submission rates of 90.8--99.1\% at $k=1$ and 85.5--98.4\% at $k=2$. Each remains above the corresponding non-thinking rate. Thus, the increase in premature submission with thinking is not explained solely by decisions one or two turns away from the annotated boundary.

\subsection{Thinking-Mode Evaluation and Belief Measurement}
\label{app:thinking-measurement}

The thinking comparison supplements \hyperref[rq:timing]{RQ1} and \hyperref[rq:belief]{RQ2} for seven models: four Qwen3 sizes, Qwen3.6-27B, Qwen3.8-27B, and DeepSeek-V4-Flash. Both modes use the same cases, candidates, evidence, temperature, and seed. With thinking enabled, the prompt permits internal reasoning but still requires a final action JSON and follows the same stopping rule.

\paragraph{Fixed-prefix measurement in both modes.}
In both modes, candidate probabilities are scored after \verb|{"action":"answer","answer":"| in the cloned context. With thinking enabled, the generated reasoning precedes this fixed prefix. An illustrative continuation is
\begin{quote}
\small\ttfamily
<think>[reasoning]</think>\par
\{"action":"answer","answer":"C"\}
\end{quote}
At the position occupied by C, we score all candidate labels A--J and normalize over those labels. The distribution is conditioned on the cloned conversation, the generated reasoning, and the fixed JSON prefix. With thinking disabled, the same procedure scores the labels without a preceding reasoning segment.

\paragraph{Candidate measurement when the model waits.}
A natural \textsc{wait} response need not contain an answer field. To measure diagnostic belief at that turn, we supply the same fixed answer-field prefix in the cloned context and score all ten candidate labels. This procedure also applies to thinking-enabled waiting turns, where the generated reasoning precedes the scoring prefix. It does not introduce a new instruction asking the model to select a diagnosis. The original \textsc{wait} action remains unchanged, so the diagnostic trajectory continues to receive evidence.

All measurements use token probabilities for the candidate alternatives rather than a numerical confidence statement generated by the model. For illustration, if label C has probability mass 0.24 and the ten labels together have mass 0.60, its normalized probability is $p(C)=0.40$. Submission and waiting checkpoints use the same scoring prefix. With thinking enabled, gold belief and ECE describe predictions conditioned on the generated reasoning.

Table~\ref{tab:thinking-sensitivity-full} uses the metrics in Appendix~\ref{app:metric-beliefs}. Early submission includes all completed trajectories, submission accuracy includes answered trajectories, and $p_{\mathrm{gold}}$ is measured at the diagnosis field during submission. Margin and ECE pool all measured evidence checkpoints. Only 3--51 cases per model reach $t_i^*$ with thinking enabled, after early submissions have ended the other trajectories.

\begin{table}[htbp]
\centering
\small
\renewcommand{\arraystretch}{1.16}
\setlength{\tabcolsep}{5mm}
\vspace{-0.5em}
\caption{Thinking-mode results. All rates, probabilities, margins, and ECE are multiplied by 100. Off/on denotes the thinking setting. $p_{\mathrm{gold}}$ is measured at diagnosis submission, whereas margin and ECE use all measured evidence checkpoints. $n_{t^*}$ counts boundary measurements with thinking on.}
\label{tab:thinking-sensitivity-full}
\vspace{0.4em}
\resizebox{\linewidth}{!}{\begin{tabular}{lrrrrrrrr}
\toprule
\rowcolor{appBhead}
 & \multicolumn{2}{c}{Early submission} & \multicolumn{2}{c}{Submission accuracy} & \multicolumn{3}{c}{Belief measurements (on)} &  \\
\cmidrule(lr){2-3}\cmidrule(lr){4-5}\cmidrule(lr){6-8}
\rowcolor{appBhead}
\textbf{Model} & Off & On & Off & On & $p_{\mathrm{gold}}$ & Margin & ECE & $n_{t^*}$ \\
\midrule
\rowcolor{appBstripe}
\textcolor{appBink}{Q3-8B} & 99.1 & 99.7 & 10.9 & 11.3 & 11.5 & 100.0 & 89.8 & 4 \\
\textcolor{appBink}{Q3-14B} & 78.6 & 99.7 & 30.5 & 12.0 & 10.4 & 99.8 & 90.1 & 5 \\
\rowcolor{appBstripe}
\textcolor{appBink}{Q3-32B} & 35.3 & 99.6 & 59.2 & 13.9 & 14.1 & 99.3 & 87.9 & 4 \\
\textcolor{appBink}{Q3-235B} & 49.7 & 99.6 & 55.4 & 17.3 & 15.9 & 99.7 & 86.5 & 3 \\
\rowcolor{appBstripe}
\textcolor{appBink}{Q3.6-27B} & 41.0 & 94.9 & 70.2 & 37.8 & 35.6 & 100.0 & 76.1 & 50 \\
\textcolor{appBink}{Q3.8-27B} & 36.6 & 94.1 & 72.9 & 36.8 & 41.0 & 100.0 & 74.4 & 51 \\
\rowcolor{appBstripe}
\textcolor{appBink}{DS-V4-F} & 24.6 & 97.3 & 70.5 & 28.5 & 16.2 & 35.7 & 39.7 & 26 \\
\bottomrule
\end{tabular}
}
\end{table}

Thinking raises premature submission to 94.1--99.7\% across the seven models. Qwen3-8B already submits prematurely on 99.1\% of cases with thinking off, and its submission accuracy rises slightly from 10.9\% to 11.3\%. Accuracy falls for the other six models. DeepSeek-V4-Flash, for example, moves from 24.6\% to 97.3\% premature submission while accuracy falls from 70.5\% to 28.5\%. All seven models show lower submission-point gold belief and higher ECE. The large probability margins in the Qwen variants therefore reflect confident preferences that are often incorrect.

\subsection{Evidence Order: Comparing Final Diagnoses and Boundary Beliefs}

\hyperref[rq:order]{RQ3} retains each original sequence and constructs up to four distinct permutations of the evidence before $t_i^*$. Using a fixed seed, we first shuffle within clinical stages and then shuffle the full prefix if needed to obtain additional distinct orders. The evidence at $t_i^*$ and afterward remains fixed. Trajectories reaching the boundary have therefore received identical evidence sets in different orders.

We compare final diagnosis outcomes for every original--permuted pair, including early-terminated trajectories. We separately compare pairs with candidate probabilities available in both orders at $t_i^*$, using each trajectory's own accumulated conversation. For these pairs, we compute boundary JSD and eventual diagnosis disagreement on the same cases. Both metrics follow the aggregation in Appendix~\ref{app:metric-interventions}, so a case with more eligible permutations receives no additional weight.

Reordering changes 29.0--52.9\% of final outcomes across the full cohort. Among pairs reaching the boundary, models other than Qwen3-8B have mean JSD values of 0.0090--0.0436 but still disagree on 6.2--20.6\% of final outcomes (Table~\ref{tab:order-cohorts}). Similar beliefs after the same evidence thus coexist with different eventual diagnoses. Qwen3-8B contributes only seven boundary-observed source cases because it usually submits earlier, producing a wide confidence interval.

\begin{table}[t]
\centering\small
\renewcommand{\arraystretch}{1.16}
\setlength{\tabcolsep}{5mm}
\vspace{-1em}
\caption{Evidence-order comparisons for pairs in which both trajectories reach $t_i^*$. Source cases and order pairs count distinct cases and eligible original--permuted comparisons, respectively. JSD and final diagnosis disagreement use the same pairs and equal source-case weights. Brackets give 95\% bootstrap confidence intervals. All order pairs from a sampled case remain together.}
\label{tab:order-cohorts}
\vspace{0.4em}
\resizebox{\linewidth}{!}{\begin{tabular}{lrrrr}
\toprule
\rowcolor{appBhead}
\textbf{Model} & Source cases & Order pairs & JSD $\times100$ & Final diagnosis disagreement (\%) \\
\midrule
\rowcolor{appBstripe}
\textcolor{appBink}{Q3-8B} & 7 & 17 & 2.12 [0.02, 5.98] & 3.6 [0.0, 10.7] \\
\textcolor{appBink}{Q3-14B} & 76 & 191 & 1.58 [0.58, 2.71] & 6.2 [1.6, 11.8] \\
\rowcolor{appBstripe}
\textcolor{appBink}{Q3-32B} & 635 & 2120 & 4.36 [3.66, 5.07] & 15.4 [13.1, 17.9] \\
\textcolor{appBink}{Q3-235B} & 353 & 955 & 2.87 [2.11, 3.77] & 13.1 [9.9, 16.4] \\
\rowcolor{appBstripe}
\textcolor{appBink}{Q3.6-27B} & 545 & 1678 & 1.28 [1.04, 1.54] & 13.8 [11.3, 16.6] \\
\textcolor{appBink}{Q3.8-27B} & 420 & 1091 & 0.90 [0.72, 1.11] & 13.2 [10.3, 16.3] \\
\rowcolor{appBstripe}
\textcolor{appBink}{DS-V4-F} & 780 & 2721 & 1.48 [1.22, 1.75] & 20.6 [18.4, 22.9] \\
\textcolor{appBink}{K2.5} & 832 & 2753 & 1.68 [1.42, 1.95] & 18.8 [16.6, 21.1] \\
\rowcolor{appBstripe}
\textcolor{appBink}{K2.6} & 807 & 2629 & 1.66 [1.36, 1.98] & 18.3 [16.0, 20.6] \\
\bottomrule
\end{tabular}
}
\vspace{-0.5em}
\end{table}

\subsection{Repeated Evidence, Irrelevant Text, and Candidate Labels}

The first evidence-value intervention in \hyperref[rq:strength]{RQ4} tests whether text without new diagnostic information changes the model's response. We select prefixes ending at approximately 25\%, 50\%, and 75\% of the evidence sequence, at $t_i^*$, and one turn before the original natural diagnosis submission when available. Prefixes contain at least one item, and coincident positions are merged. At each position, two branches share the same evidence prefix. One repeats its first, middle, or most recent evidence item. The other adds a clinically irrelevant statement, such as ``The weather is sunny today.'' Neither supplies a new clinical fact that helps distinguish among the candidate diagnoses.

To measure the response at each selected position, we reconstruct the prefix in a separate conversation, recording \textsc{wait} after each preceding evidence item. We then append the repeated evidence or irrelevant statement and let the model choose its next action. We measure beliefs before and after the addition and compare the diagnosis outputs of the two branches immediately afterward. A difference occurs if the model submits different diagnoses or submits a diagnosis in one branch while waiting in the other. These are local responses to the inserted text at a common evidence prefix.

\begin{figure}[htbp]
  \centering
  \includegraphics[width=\linewidth]{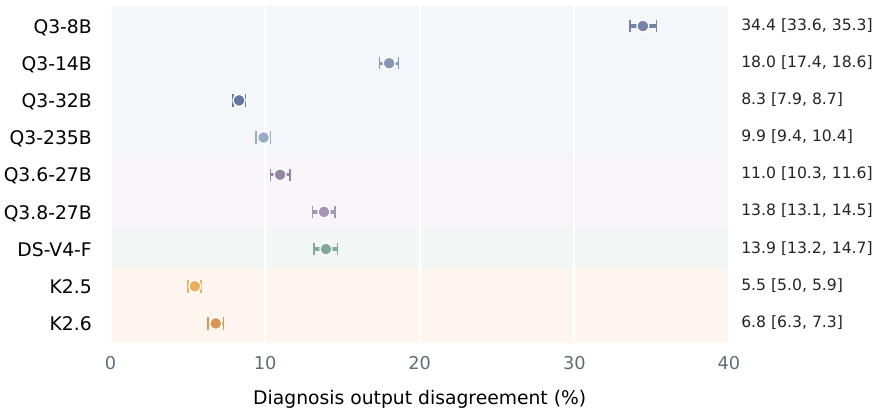}
  \caption{Diagnosis output differences immediately after repeated evidence or irrelevant text is added to the same evidence prefix. Points show the percentage of comparisons with different submitted diagnoses or a diagnosis in only one branch. Intervals are source-case bootstrap 95\% confidence intervals, with comparisons averaged within each case.}
  \label{fig:valueless-evidence}
\end{figure}

Diagnosis outputs differ in 5.5--34.4\% of comparisons (Figure~\ref{fig:valueless-evidence}), with the largest effect in Qwen3-8B. Although neither addition supplies new diagnostic information, the choice of text changes whether the model submits a diagnosis and which diagnosis it submits.

A separate control tests candidate-label sensitivity. For each case, four fixed-seed permutations reassign the ten letters to the existing diagnosis rows while preserving diagnosis text, row order, and evidence. For example, a diagnosis originally labeled A may receive F in the same row. We run each version naturally and map submitted labels back to diagnoses before comparison. Final outcomes differ from the original-label run in 9.7--28.8\% of cases, including diagnosis-versus-non-answer differences. The identifiers alone can therefore influence the final outcome.

\subsection{Misleading Evidence: Construction, Measurements, and Results}

The strength intervention in \hyperref[rq:strength]{RQ4} examines how evidence favoring an incorrect diagnosis changes beliefs and subsequent submissions. We first synthesize misleading evidence for the 1,050 cases in \OURS{}. For each case, an incorrect candidate serves as the target diagnosis. We construct weak and strong observations that favor this target while matching their clinical event type, semantic direction, and presentation format. Each version is inserted early in the sequence, immediately before the original sufficient-evidence item, or immediately after it. The source case, target, and insertion position remain fixed within a strong--weak pair.

Before forming the test set, Claude Opus 4.8, Gemini 3.1 Pro Preview, and GPT-5.6-sol independently review the augmented evidence sequences. The reviewers judge whether the original decisive evidence still marks the earliest point at which the correct diagnosis can be distinguished from the candidates. When insertion precedes that item, its turn index shifts forward by one. All three reviewers agree that the boundary remains valid for 553 cases, from which we manually select 318 for the intervention experiment. The construction review also checks that inserted findings are clinically coherent, temporally possible, and free of information from later evidence.

\paragraph{Illustrative boundary checks.}
The following two high-strength examples show the complete evidence available through the original sufficient-evidence item, together with the inserted distractor. Both distractors support a competing diagnosis without replacing the source-report evidence that identifies the correct diagnosis.

In the first case, hearing loss, tympanic changes, and the inserted flat tympanogram make middle-ear effusion plausible before evidence E9. The histopathologic finding of branched fungal hyphae is the first observation that identifies the fungus ball among the candidates. The insertion therefore changes the turn number of that finding from 9 to 10 without changing the correct diagnosis or the decisive evidence. In the second case, cervical inflammation may coexist with the presenting symptoms, but it arrives after the hysterosalpingogram has shown contrast entering the bowel. It does not explain that finding or move the earliest sufficient turn away from evidence E15.

\begingroup
\definecolor{misleadboxbg}{RGB}{247,249,252}
\definecolor{misleadboxline}{RGB}{151,174,196}
\definecolor{misleadboxhead}{RGB}{55,87,119}
\setlength{\fboxsep}{8pt}
\setlength{\fboxrule}{0.6pt}
\newcommand{\misleadevidence}[2]{\par\noindent{\color{misleadboxhead}\textbf{#1}}\enspace #2}
\par\smallskip\noindent
\fcolorbox{misleadboxline}{misleadboxbg}{%
\begin{minipage}{\dimexpr\linewidth-2\fboxsep-2\fboxrule\relax}
\footnotesize\raggedright
{\color{misleadboxhead}\textbf{Before the boundary: middle-ear fungus ball}}\hfill\texttt{33872bd24da49e2633decfbe0a4d0186}\par\smallskip
\textbf{Correct diagnosis:} middle-ear fungus ball. \textbf{Incorrect target:} otitis media with effusion due to Eustachian tube dysfunction.\par\smallskip
\textbf{Evidence through the original $t^*=9$ (E1--E9):}
\misleadevidence{E1}{The patient was a 56-year-old Thai man.}
\misleadevidence{E2}{He presented with gradually increasing hearing loss in the right ear for 1 year.}
\misleadevidence{E3}{He had undergone right tympanoplasty for chronic suppurative otitis media 21 years earlier, resulting in cure with good hearing.}
\misleadevidence{E4}{Otoscopic examination showed localized retraction of the tympanic membrane in the posterosuperior quadrant.}
\misleadevidence{E5}{The remainder of the tympanic membrane was dull and thick.}
\misleadevidence{E6}{Audiological examination showed conductive hearing loss with a 55 dB speech reception threshold, 40 dB bone conduction level, and 30 dB air-bone gap.}
\misleadevidence{E7}{Exploratory tympanotomy revealed a muddy clay-like substance in the middle-ear cavity mixed with thick brown fluid, along with debris and granulation.}
\misleadevidence{E8}{The ossicular chain appeared intact during exploratory tympanotomy.}
\misleadevidence{Insert}{The right-ear tympanogram shows a flat type B pattern with normal ear-canal volume, supporting middle-ear fluid accumulation from impaired ventilation.}
\misleadevidence{E9}{Histopathology showed a tangled mass of branched hyphae surrounded by numerous neutrophils.}
\par\smallskip\textbf{Boundary after insertion:} original E9 appears at augmented turn 10.
\end{minipage}}%
\par\smallskip\noindent
\fcolorbox{misleadboxline}{misleadboxbg}{%
\begin{minipage}{\dimexpr\linewidth-2\fboxsep-2\fboxrule\relax}
\footnotesize\raggedright
{\color{misleadboxhead}\textbf{After the boundary: salpingo-enteric fistula}}\hfill\texttt{085b81c229b8accdee45113642cb7959}\par\smallskip
\textbf{Correct diagnosis:} salpingo-enteric fistula. \textbf{Incorrect target:} chronic cervicitis.\par\smallskip
\textbf{Evidence through the original $t^*=15$ (E1--E15):}
\misleadevidence{E1}{The patient was a 34-year-old Para 0+1 woman.}
\misleadevidence{E2}{She presented for routine hysterosalpingographic evaluation of secondary infertility after eight years of marriage without achieving pregnancy.}
\misleadevidence{E3}{She reported intermittent lower abdominal pain.}
\misleadevidence{E4}{She reported occasional intermittent vaginal discharge.}
\misleadevidence{E5}{She had a history of chronic pelvic inflammatory disease.}
\misleadevidence{E6}{She had an uneventful termination of a seven-week pregnancy two years before marriage.}
\misleadevidence{E7}{She underwent myomectomy three years before presentation.}
\misleadevidence{E8}{Hysterosalpingography performed before the myomectomy showed fallopian tubes of normal caliber that were patent bilaterally.}
\misleadevidence{E9}{She underwent appendicectomy approximately 14 months before presentation.}
\misleadevidence{E10}{She had completed oral ciprofloxacin and metronidazole therapy.}
\misleadevidence{E11}{Physical examination showed copious vaginal discharge.}
\misleadevidence{E12}{Initial hysterosalpingography opacified a moderate-capacity uterus with regular margins.}
\misleadevidence{E13}{The left fallopian tube did not opacify on hysterosalpingography.}
\misleadevidence{E14}{The right fallopian tube was visualized with a triangular loculation of contrast at its distal aspect.}
\misleadevidence{E15}{Subsequent hysterosalpingographic films showed contrast opacification of the large bowel, with haustrations visible within the opacified bowel.}
\misleadevidence{Insert}{The cervix has marked erythema, friability, and mucopurulent endocervical discharge.}
\par\smallskip\textbf{Boundary after insertion:} original E15 remains at augmented turn 15.
\end{minipage}}%
\par\smallskip
\endgroup

\paragraph{First, measure beliefs at specified evidence positions.}
For each intervention version, we construct separate conversations containing the candidate list and accumulated evidence up to four positions: immediately before insertion, immediately after insertion, upon receipt of the original sufficient-evidence item, and at the end of the complete sequence. These measurement conversations contain no intervening diagnosis actions. This allows us to examine all four evidence positions for every intervention version, even when a natural trajectory would answer earlier. For insertion after the boundary, the boundary measurement precedes insertion. The positions are evaluated in their actual sequence order, preserving the clinical history available at each measurement.

At each position, we record $p_{\mathrm{target}}$, $p_{\mathrm{gold}}$, their difference, and the highest-probability candidate, together with submission propensity $p(\mathrm{answer})$. The belief metrics and strong-minus-weak contrasts are defined in Appendix~\ref{app:metric-interventions}. Measurements immediately before and after insertion quantify the initial response, while the final prefix measures the effect remaining after all original evidence has arrived, including the evidence that establishes the correct diagnosis.

\paragraph{Second, evaluate natural diagnosis submissions.}
Each strong and weak version also runs as a separate natural trajectory, with the model free to wait or submit after each item. The first submission ends the interaction. We record target submissions, submission timing, correctness, and non-answers. These outcomes link the measured belief shifts to actual diagnosis submissions, including failures to submit after the complete evidence sequence has been revealed.

\paragraph{Third, summarize the effects of stronger misleading evidence.}
Table~\ref{tab:strength-effects} gives the five effects defined in Appendix~\ref{app:metric-interventions}, following the column order of Figure~\ref{fig:strength-cascade}a. The first four use early insertion: the immediate target-belief increase, target-submission increase, diagnostic accuracy loss, and final target-belief increase. The fifth uses insertion after the original boundary and reports final correct-top-1 loss. All effects are expressed in percentage points, with positive values indicating harm.

\begin{table}[htbp]
\centering\small
\renewcommand{\arraystretch}{1.18}
\setlength{\tabcolsep}{3.5mm}
\caption{Effects of stronger misleading evidence across models (percentage points). The first four endpoints use early insertion, and the last uses insertion after $t_i^*$. All positive values indicate harm. Brackets give source-case bootstrap 95\% confidence intervals. Final diagnostic accuracy counts non-answers as errors. Bold values mark the smallest observed harmful effect in each column.}
\label{tab:strength-effects}
\vspace{0.4em}
\resizebox{\linewidth}{!}{\begin{tabular}{lccccc}
\toprule
\rowcolor{appBhead}
 & \multicolumn{4}{c}{Early insertion} & After $t^*$ \\
\cmidrule(lr){2-5}\cmidrule(lr){6-6}
\rowcolor{appBhead}
\textbf{Model} & \shortstack{Target belief\\at insertion $\uparrow$} & \shortstack{Target diagnosis\\submission $\uparrow$} & \shortstack{Diagnostic\\accuracy $\downarrow$} & \shortstack{Target belief\\at completion $\uparrow$} & \shortstack{Correct diagnosis\\ranked first $\downarrow$} \\
\midrule
\rowcolor{appBstripe}
\textcolor{appBink}{Q3-8B} & \shortstack{\textbf{11.1}\\\scriptsize \textcolor{appBmuted}{[9.3, 13.1]}} & \shortstack{\textbf{3.0}\\\scriptsize \textcolor{appBmuted}{[1.7, 4.5]}} & \shortstack{\textbf{1.0}\\\scriptsize \textcolor{appBmuted}{[0.2, 1.9]}} & \shortstack{\textbf{4.4}\\\scriptsize \textcolor{appBmuted}{[3.4, 5.5]}} & \shortstack{\textbf{3.9}\\\scriptsize \textcolor{appBmuted}{[2.2, 5.8]}} \\[2pt]
\textcolor{appBink}{Q3-14B} & \shortstack{18.2\\\scriptsize \textcolor{appBmuted}{[15.9, 20.6]}} & \shortstack{21.7\\\scriptsize \textcolor{appBmuted}{[18.6, 24.8]}} & \shortstack{7.2\\\scriptsize \textcolor{appBmuted}{[4.9, 9.7]}} & \shortstack{9.1\\\scriptsize \textcolor{appBmuted}{[7.4, 10.8]}} & \shortstack{6.0\\\scriptsize \textcolor{appBmuted}{[4.1, 8.0]}} \\[2pt]
\rowcolor{appBstripe}
\textcolor{appBink}{Q3-32B} & \shortstack{17.7\\\scriptsize \textcolor{appBmuted}{[15.7, 19.7]}} & \shortstack{24.0\\\scriptsize \textcolor{appBmuted}{[21.2, 27.0]}} & \shortstack{12.3\\\scriptsize \textcolor{appBmuted}{[9.6, 15.1]}} & \shortstack{10.8\\\scriptsize \textcolor{appBmuted}{[9.1, 12.6]}} & \shortstack{11.1\\\scriptsize \textcolor{appBmuted}{[8.7, 13.5]}} \\[2pt]
\textcolor{appBink}{Q3-235B} & \shortstack{18.4\\\scriptsize \textcolor{appBmuted}{[16.0, 20.8]}} & \shortstack{33.2\\\scriptsize \textcolor{appBmuted}{[30.1, 36.4]}} & \shortstack{19.6\\\scriptsize \textcolor{appBmuted}{[16.5, 22.7]}} & \shortstack{9.9\\\scriptsize \textcolor{appBmuted}{[8.3, 11.8]}} & \shortstack{11.9\\\scriptsize \textcolor{appBmuted}{[9.3, 14.5]}} \\[2pt]
\rowcolor{appBstripe}
\textcolor{appBink}{Q3.6-27B} & \shortstack{24.6\\\scriptsize \textcolor{appBmuted}{[22.8, 26.3]}} & \shortstack{35.0\\\scriptsize \textcolor{appBmuted}{[31.8, 38.5]}} & \shortstack{26.7\\\scriptsize \textcolor{appBmuted}{[23.1, 30.0]}} & \shortstack{8.6\\\scriptsize \textcolor{appBmuted}{[7.4, 9.9]}} & \shortstack{16.6\\\scriptsize \textcolor{appBmuted}{[14.1, 19.2]}} \\[2pt]
\textcolor{appBink}{Q3.8-27B} & \shortstack{20.3\\\scriptsize \textcolor{appBmuted}{[18.7, 21.9]}} & \shortstack{34.2\\\scriptsize \textcolor{appBmuted}{[31.2, 37.4]}} & \shortstack{25.1\\\scriptsize \textcolor{appBmuted}{[21.8, 28.5]}} & \shortstack{6.1\\\scriptsize \textcolor{appBmuted}{[5.1, 7.2]}} & \shortstack{12.2\\\scriptsize \textcolor{appBmuted}{[9.8, 14.9]}} \\[2pt]
\rowcolor{appBstripe}
\textcolor{appBink}{DS-V4-F} & \shortstack{20.2\\\scriptsize \textcolor{appBmuted}{[18.2, 22.2]}} & \shortstack{29.6\\\scriptsize \textcolor{appBmuted}{[26.4, 32.9]}} & \shortstack{10.4\\\scriptsize \textcolor{appBmuted}{[7.5, 13.2]}} & \shortstack{7.7\\\scriptsize \textcolor{appBmuted}{[6.1, 9.4]}} & \shortstack{14.1\\\scriptsize \textcolor{appBmuted}{[11.5, 16.8]}} \\[2pt]
\textcolor{appBink}{K2.5} & \shortstack{21.1\\\scriptsize \textcolor{appBmuted}{[19.1, 23.1]}} & \shortstack{33.5\\\scriptsize \textcolor{appBmuted}{[30.3, 36.7]}} & \shortstack{18.6\\\scriptsize \textcolor{appBmuted}{[15.4, 21.6]}} & \shortstack{10.8\\\scriptsize \textcolor{appBmuted}{[9.3, 12.4]}} & \shortstack{17.9\\\scriptsize \textcolor{appBmuted}{[15.2, 20.5]}} \\[2pt]
\rowcolor{appBstripe}
\textcolor{appBink}{K2.6} & \shortstack{24.3\\\scriptsize \textcolor{appBmuted}{[22.2, 26.3]}} & \shortstack{31.4\\\scriptsize \textcolor{appBmuted}{[28.4, 34.7]}} & \shortstack{16.9\\\scriptsize \textcolor{appBmuted}{[13.9, 20.1]}} & \shortstack{12.5\\\scriptsize \textcolor{appBmuted}{[10.8, 14.2]}} & \shortstack{15.3\\\scriptsize \textcolor{appBmuted}{[12.6, 17.9]}} \\[2pt]
\bottomrule
\end{tabular}
}
\end{table}

All five effects are positive in every model, with confidence intervals excluding zero. Stronger evidence adds 11.1--24.6 points to the immediate increase in target belief. For models other than Qwen3-8B, target submission rises by 21.7--35.0 points and diagnostic accuracy falls by 7.2--26.7 points. After all original evidence arrives, target belief remains 4.4--12.5 points higher. Insertion after the original boundary reduces final correct-top-1 frequency by 3.9--17.9 points. The influence of stronger misleading evidence therefore persists in both beliefs and submitted diagnoses.

Qwen3-8B shows smaller behavioral changes: target submission increases by 3.0 points and accuracy falls by 1.0 point, despite an 11.1-point additional belief shift. It submits very early under both strengths, leaving fewer trajectories whose outcomes can change further.

\paragraph{Interpreting insertion before and after the boundary.}
Within each insertion position, strong and weak evidence are compared against the same clinical history. The median additional target-belief increase across models is 20.2 points for early insertion, 24.1 immediately before the boundary, and 26.1 immediately afterward (Figure~\ref{fig:strength-cascade}b). Beliefs remain responsive to misleading evidence after the original evidence becomes sufficient. Each insertion position has a different evidence history.

\section{Additional Diagnostic-Impact Analyses}
\label{app:impact-details}

\definecolor{appCink}{RGB}{50,78,112}
\definecolor{appCmuted}{RGB}{104,116,131}

This appendix separates the prediction of diagnosis submission from the assessment of errors at submission, then details the Natural Gate intervention in Section~\ref{sec:impact}. It also examines diagnostic beliefs after prematurely stopped cases receive sufficient evidence. Metric definitions appear in Appendix~\ref{app:metric-definitions}.

\subsection{Separating Submission Prediction from Error Assessment}
\label{app:impact-prediction-protocol}

\paragraph{Two prediction targets.}
We use the existing non-thinking trajectories and probability measurements for all nine LLMs from Sections~\ref{sec:benchmark} and~\ref{sec:mechanism}. No additional diagnostic trajectories are generated. The first task asks whether a model that currently waits will submit a diagnosis on the next turn. The second asks whether a diagnosis is incorrect when the model submits it. Separating these targets allows us to test whether EVM features carry information about diagnostic correctness in addition to their ability to predict submission behavior.

For next-turn submission prediction, an eligible example is a checkpoint at which the model chooses \textsc{wait} and a subsequent turn is observed. Inputs are measured through the current turn $t$, and the label is one when the action at $t+1$ is \textsc{answer}, regardless of which diagnosis is submitted. A further \textsc{wait} gives label zero. For example, waiting at turn 8 and submitting at turn 9 produces a positive example at turn 8, whether the diagnosis is correct or incorrect. Before balancing, this support contains 163,551 checkpoints and 7,941 positive examples. The submission frequency varies from 3.1\% to 20.0\% across LLMs, motivating the class-balanced comparison below.

For error assessment at submission, each trajectory contributes its first diagnosis submission, yielding 7,942 examples before balancing. Inputs are measured at that submission turn, and the label is one if the submitted diagnosis differs from the ground truth. Correct submissions have label zero. Cases without a submitted diagnosis do not enter this task. Thus, this experiment evaluates the ability to distinguish incorrect from correct submissions using the state at submission. The preceding experiment separately measures prediction of future behavior from a waiting checkpoint.

\paragraph{Shared predictor inputs.}
The feature definitions are the same for both tasks, with measurements taken at their respective prediction checkpoints. The progress baseline M0 uses the fraction of evidence received, $t/T_i$, the total evidence count $T_i$, and the number of candidates. M1 adds diagnostic confidence: the highest candidate probability, the difference between the two highest probabilities, and the entropy of the candidate distribution. These quantities summarize the certainty and concentration of the diagnostic belief distribution without identifying the correct candidate.

M2 supplements M0 with the original EVM features. These describe diagnosis submission propensity, its relationship to diagnostic confidence, and changes in beliefs across consecutive turns. Submission propensity is the probability of choosing \textsc{answer} rather than \textsc{wait} under the action probe. Its turn-to-turn change captures whether the model becomes more inclined to submit as new evidence arrives. The remaining features compare submission propensity with confidence and entropy and summarize fluctuations in confidence, probability margin, and entropy. At the first turn, change features are set to zero. M3 combines the inputs of M1 and M2. Neither M1 nor M3 uses the sufficiency annotation, submission lead time, or the identity of the correct diagnosis.

For next-turn submission, $M_{\mathrm{prop}}$ uses only submission propensity to examine how much predictive information is already present in the action probability. For submission errors, we additionally report $M_{\mathrm{timing}}$, which adds the annotated lead time $t_i^*-\tau_i$ to M0, where $\tau_i$ is the first submission turn. An oracle reference adds the probability of the ground-truth diagnosis to M1. These two references quantify the information provided by the annotated boundary and answer key, respectively. Their privileged inputs are separate from the M1--M3 comparison.

For completeness, let $u_t$ be submission propensity, $c_t$ the highest candidate probability, $m_t$ the margin between the two highest probabilities, and $H_t$ the candidate entropy, suppressing the case index. With $\Delta x_t=x_t-x_{t-1}$, the six EVM features are
\begin{equation}
\left(u_t,\ u_tm_t,\ \Delta u_t,\ u_t(1-c_t),\ u_tH_t,\
|\Delta c_t|+|\Delta m_t|+|\Delta H_t|\right).
\end{equation}
The products describe how the inclination to submit interacts with diagnostic certainty or uncertainty. The final term summarizes how much the belief distribution changes as evidence arrives. These features have the same definitions at waiting checkpoints and submission turns, allowing the two tasks to evaluate the same information under different targets.

\paragraph{Case splits and class balancing.}
Case identifiers are assigned deterministically to 20\% training, 20\% validation, and 60\% test partitions using seed 2027. All checkpoints belonging to a case remain in the same partition, and the assignment is shared across predictor specifications. Within each partition and task, we then undersample the majority class to match the minority class using the same seed. Examples are neither duplicated nor moved between partitions. Each retained partition therefore contains equal numbers of positive and negative examples, and competing predictors are evaluated on the same retained examples.

Balancing is applied separately to submission versus waiting in the first task and incorrect versus correct submissions in the second. This prevents the majority class from dominating the reported probability losses and provides a common class proportion across models. The reported log losses consequently evaluate probabilities under this balanced sampling distribution. The original checkpoint counts describe the available support before undersampling, rather than the size of the balanced test sets.

\paragraph{Regression fitting and evaluation.}
We fit a separate L2-regularized logistic regression for each LLM, prediction task, and feature specification. Continuous inputs are standardized using training-partition statistics. The penalty is selected from $\{0.01,0.1,1,10\}$ by validation log loss, after which the predictor is refit using the training and validation partitions with the original standardization. Testing uses held-out cases. Pooled regressions combine the LLMs and add model-identity indicators while retaining the case-level split.

AUROC evaluates how well predicted scores distinguish positive from negative examples. For next-turn submission, positives are upcoming submissions. For error assessment, positives are incorrect submitted diagnoses. Log loss evaluates the probability assigned to the realized binary outcome, as defined in Equation~\ref{eq:log-loss}. For next-turn submission, we report $\Delta\mathrm{LL}=\mathrm{LL}_{M3}-\mathrm{LL}_{M1}$, where negative values favor M3. The balanced evaluation reports held-out point estimates for both prediction tasks.

\begin{table}[htbp]
\centering\small
\renewcommand{\arraystretch}{1.12}
\setlength{\tabcolsep}{5mm}
\caption{Balanced next-turn submission prediction. AUROC compares upcoming diagnosis submissions with continued waiting. M3 combines confidence and EVM features, and $\Delta$ is its improvement over M1.}
\label{tab:impact-prediction-full}
\vspace{0.4em}
\resizebox{\linewidth}{!}{\begin{tabular}{lrrrrrr}
\toprule
Model & M0 & M1 & M2 & $M_{\mathrm{prop}}$ & M3 & $\Delta$(M3$-$M1) \\
\midrule
Qwen3-8B & 0.791 & 0.795 & 0.812 & 0.838 & 0.820 & +0.025 \\
Qwen3-14B & 0.778 & 0.780 & 0.806 & 0.824 & 0.827 & +0.047 \\
Qwen3-32B & 0.830 & 0.826 & 0.846 & 0.806 & 0.846 & +0.020 \\
Qwen3-235B-A22B & 0.831 & 0.827 & 0.844 & 0.772 & 0.840 & +0.013 \\
Qwen3.6-27B & 0.813 & 0.815 & 0.821 & 0.765 & 0.812 & -0.003 \\
Qwen3.8-27B & 0.850 & 0.857 & 0.877 & 0.843 & 0.885 & +0.028 \\
Kimi-K2.5 & 0.850 & 0.851 & 0.855 & 0.750 & 0.842 & -0.009 \\
Kimi-K2.6 & 0.868 & 0.863 & 0.877 & 0.771 & 0.873 & +0.010 \\
DeepSeek-V4-Flash & 0.792 & 0.790 & 0.812 & 0.730 & 0.807 & +0.017 \\
\midrule
Pooled & 0.833 & 0.835 & 0.849 & 0.735 & 0.854 & +0.019 \\
\bottomrule
\end{tabular}
}
\end{table}

\paragraph{Predicting submission behavior.}
M3 improves AUROC over M1 for seven of nine LLMs, with pooled AUROC increasing from 0.835 to 0.854 (Table~\ref{tab:impact-prediction-full}). Improvements are not uniform. Qwen3.6-27B and Kimi-K2.5 show small declines, while submission propensity alone is competitive for Qwen3-8B. These comparisons indicate that the relation between action probabilities, changing beliefs, and subsequent submission differs across models. Combining these signals is generally useful, but it does not consistently dominate every simpler predictor.

\begin{table}[!ht]
\centering\small
\renewcommand{\arraystretch}{1.12}
\setlength{\tabcolsep}{5mm}
\caption{Log loss for balanced next-turn submission prediction. Lower values indicate better submission-probability estimates, and negative M3$-$M1 differences favor the addition of EVM features.}
\label{tab:impact-prediction-ll}
\vspace{0.4em}
\resizebox{\linewidth}{!}{\begin{tabular}{lrrrrrr}
\toprule
Model & M0 & M1 & M2 & $M_{\mathrm{prop}}$ & M3 & $\Delta$(M3$-$M1) \\
\midrule
Qwen3-8B & 0.4369 & 0.4340 & 0.4179 & 0.4575 & 0.3985 & -0.0355 \\
Qwen3-14B & 0.2175 & 0.2166 & 0.2104 & 0.2395 & 0.2041 & -0.0125 \\
Qwen3-32B & 0.1428 & 0.1434 & 0.1392 & 0.1653 & 0.1391 & -0.0043 \\
Qwen3-235B-A22B & 0.1682 & 0.1693 & 0.1645 & 0.1987 & 0.1635 & -0.0058 \\
Qwen3.6-27B & 0.1620 & 0.1611 & 0.1600 & 0.1837 & 0.1600 & -0.0011 \\
Qwen3.8-27B & 0.1472 & 0.1453 & 0.1418 & 0.1729 & 0.1378 & -0.0075 \\
Kimi-K2.5 & 0.1135 & 0.1139 & 0.1139 & 0.1378 & 0.1145 & +0.0006 \\
Kimi-K2.6 & 0.1194 & 0.1194 & 0.1185 & 0.1426 & 0.1186 & -0.0008 \\
DeepSeek-V4-Flash & 0.1195 & 0.1193 & 0.1180 & 0.1352 & 0.1163 & -0.0030 \\
\midrule
Pooled & 0.1562 & 0.1556 & 0.1527 & 0.1780 & 0.1494 & -0.0062 \\
\bottomrule
\end{tabular}
}
\end{table}

The probability estimates also improve for most models. M3 reduces log loss relative to M1 for eight LLMs, and pooled log loss falls from 0.1556 to 0.1494 (Table~\ref{tab:impact-prediction-ll}). Kimi-K2.5 is the exception on both metrics. The overall pattern supports the use of EVM features to anticipate submission behavior beyond evidence progress and candidate confidence.

\begin{table}[htbp]
\centering\small
\renewcommand{\arraystretch}{1.12}
\setlength{\tabcolsep}{4mm}
\caption{Balanced assessment of incorrect diagnoses at submission, measured by AUROC. M1 and M3 use current and preceding model measurements. The timing and oracle references additionally use the annotated boundary and correct-answer identity, respectively.}
\label{tab:impact-submission-error}
\vspace{0.4em}
\resizebox{\linewidth}{!}{\begin{tabular}{lrrrrrrr}
\toprule
Model & M0 & M1 & M2 & $M_{\mathrm{timing}}$ & M3 & Oracle & $\Delta$(M3$-$M1) \\
\midrule
Qwen3-8B & 0.625 & 0.640 & 0.637 & 0.618 & 0.724 & 1.000 & +0.084 \\
Qwen3-14B & 0.659 & 0.627 & 0.635 & 0.731 & 0.822 & 1.000 & +0.195 \\
Qwen3-32B & 0.665 & 0.711 & 0.724 & 0.750 & 0.841 & 1.000 & +0.130 \\
Qwen3-235B-A22B & 0.674 & 0.789 & 0.788 & 0.773 & 0.847 & 1.000 & +0.058 \\
Qwen3.6-27B & 0.698 & 0.783 & 0.790 & 0.809 & 0.777 & 0.999 & -0.006 \\
Qwen3.8-27B & 0.630 & 0.773 & 0.787 & 0.751 & 0.789 & 1.000 & +0.016 \\
Kimi-K2.5 & 0.708 & 0.800 & 0.821 & 0.812 & 0.800 & 1.000 & +0.000 \\
Kimi-K2.6 & 0.675 & 0.711 & 0.726 & 0.782 & 0.723 & 0.998 & +0.012 \\
DeepSeek-V4-Flash & 0.719 & 0.814 & 0.813 & 0.816 & 0.839 & 0.986 & +0.025 \\
\midrule
Pooled & 0.807 & 0.845 & 0.851 & 0.855 & 0.879 & 1.000 & +0.034 \\
\bottomrule
\end{tabular}
}
\end{table}

\FloatBarrier
\paragraph{Assessing errors at submission.}
M3 improves AUROC for seven LLMs, matches M1 for Kimi-K2.5 at the reported precision, and slightly reduces AUROC for Qwen3.6-27B (Table~\ref{tab:impact-submission-error}). Pooled AUROC increases from 0.845 to 0.879. The largest gains occur for Qwen3-14B and Qwen3-32B, showing that submission-related information can distinguish incorrect from correct diagnoses even after restricting the support to actual submissions.

The timing reference is informative, but its ranking relative to M3 varies across models. The answer-key oracle achieves near-perfect discrimination, indicating that knowing which candidate is correct supplies information far beyond ordinary confidence statistics. Neither reference is needed for the reported M3 gains. Together, the two tasks support distinct conclusions: EVM features help anticipate when a model will submit, and they help assess correctness at the point of submission. The Natural Gate experiment below then tests whether changing submission behavior can improve diagnosis outcomes.

\subsection{Diagnostic Beliefs After Receiving Sufficient Evidence}
\label{app:impact-boundary-rescue}

This experiment asks whether a model that submits a diagnosis prematurely can identify the correct diagnosis once the evidence reaches the annotated sufficiency boundary. It connects premature stopping to the information the model would otherwise miss and motivates the Natural Gate intervention below.

For each LLM, we select cases whose original natural trajectory ends with a diagnosis submission at turn $s_i<t_i^*$. Here, $t_i^*$ is the earliest annotated turn at which the accumulated evidence is sufficient to identify the correct diagnosis among the ten candidates. For each selected case, we construct three evidence prefixes in the original order: the prefix ending at the original submission turn $s_i$, the prefix ending immediately before the boundary at $t_i^*-1$, and the prefix ending at the boundary $t_i^*$. If the original submission occurs at $t_i^*-1$, the first two prefixes coincide.

Each prefix is evaluated in a separate context containing the candidate list and the evidence through that checkpoint. The original submitted diagnosis is not included in these contexts. We measure the candidate probability distribution using the same candidate-scoring procedure as in Appendix~\ref{app:mechanism-protocol}, and record both the highest-probability diagnosis and $p_{\mathrm{gold}}$, the probability assigned to the correct diagnosis. The correct answer is used only to score the resulting distribution. These measurements assess the model's diagnostic beliefs at each evidence prefix. The following Natural Gate experiment separately measures the diagnosis the model actually submits when allowed to continue.

We compare the two adjacent prefixes at $t_i^*-1$ and $t_i^*$ to measure the change associated with receiving the evidence item that reaches the sufficiency boundary. Across the nine LLMs, the mean increase in $p_{\mathrm{gold}}$ is 0.245--0.333. Among cases with an incorrect premature submission in the original trajectory, 49.2--78.4\% assign the highest probability to the correct diagnosis after receiving evidence through $t_i^*$. Thus, for a substantial fraction of premature errors, the model can recognize the correct diagnosis from evidence that its original stopping decision prevented it from seeing.

We also examine the subset whose original submission occurs exactly at $t_i^*-1$. The fraction with a correct highest-probability diagnosis at $t_i^*$ exceeds the original diagnosis accuracy by 19.7--33.3 percentage points across models. This comparison makes the missed opportunity concrete: even when the next evidence item is sufficient to resolve the case, premature submission can end the trajectory immediately before it arrives.

\subsection{Natural Gate: Delaying Submission Until Evidence Is Sufficient}
\label{app:impact-natural-gate}

\paragraph{Purpose and case selection.}
Natural Gate tests whether allowing models to receive sufficient evidence improves their final diagnosis outcomes, as reported in Section~\ref{sec:impact}. For each LLM, the intervention includes every case in which the original natural run submits a diagnosis before $t_i^*$. Selection depends on submission timing, so it includes both correct and incorrect premature diagnoses. Each selected case contributes a pair of outcomes: its original diagnosis outcome and its outcome under the gate. The candidate diagnoses, correct answer, evidence sequence, and generation configuration are held fixed within each pair.

\paragraph{How the gate operates.}
We rerun each selected case from the beginning under the natural interaction protocol. After each evidence item, the model produces its usual \textsc{wait} or \textsc{answer} action. Before $t_i^*$, an attempted diagnosis submission is blocked from terminating the case, and the next evidence item is delivered. The blocked diagnosis is omitted from the subsequent conversation history, while ordinary \textsc{wait} responses are retained. In this way, the gate implements continued waiting without feeding the blocked diagnosis back to the model.

Once the evidence item at $t_i^*$ has been delivered, the gate is removed. From that turn onward, the model can either submit a diagnosis or continue waiting under the original protocol. The first subsequent diagnosis submission ends the trajectory. If the model continues waiting through the complete evidence sequence, the case ends without a submitted diagnosis. For example, if $t_i^*=12$, attempted submissions at turns 7 and 10 are both blocked. At turn 12, the model receives the twelfth evidence item and is free to submit a diagnosis or request further evidence.

The release turn uses the offline sufficiency annotation from Appendix~\ref{app:benchmark-construction}. The model receives neither the correct diagnosis nor an instruction to submit at that turn. Access to the annotated boundary allows the experiment to measure the effect of preventing premature termination.

\paragraph{Reported outcomes.}
Table~\ref{tab:impact-gate-full} uses the definitions in Appendix~\ref{app:metric-impact}. The upper panel gives the number of selected cases, original and gated accuracy, gated submission rate, and paired accuracy gain. Every rate in this panel uses all selected cases as the denominator, including gated non-answers. The lower panel reports errors corrected, correct diagnoses lost, the confidence interval for their net effect, and the resulting gain over all 1,050 benchmark cases. Nonselected cases retain their original outcomes in this last calculation.

\begin{table}[!ht]
\centering
\small
\renewcommand{\arraystretch}{1.18}
\setlength{\tabcolsep}{5pt}
\caption{Natural Gate outcomes for each LLM. The upper panel reports accuracy and submission rates on cases with an original premature diagnosis. The lower panel decomposes the paired accuracy change into errors corrected and correct diagnoses lost, and shows the corresponding gain over all 1,050 cases. Rates are percentages, and gains and confidence limits are percentage points (pp). Bold values highlight gated accuracy and the net accuracy gain on selected cases.}
\label{tab:impact-gate-full}
\vspace{0.4em}
\resizebox{\linewidth}{!}{\begin{tabular}{lrrrrr}
\toprule
\multicolumn{6}{l}{\textcolor{appCink}{\textbf{Diagnosis outcomes on selected cases}}} \\
\cmidrule(lr){1-6}\addlinespace[3pt]
\textbf{Model} & \textbf{Cases} & \shortstack{\textbf{Original}\\\textbf{accuracy (\%)}} & \shortstack{\textbf{Gated}\\\textbf{accuracy (\%)}} & \shortstack{\textbf{Submission}\\\textbf{rate (\%)}} & \shortstack{\textbf{Accuracy}\\\textbf{gain (pp)}} \\
\midrule
\textcolor{appCink}{Qwen3-8B} & 1,041 & 10.4 & \textbf{55.6} & 100.0 & \textbf{+45.2} \\
\textcolor{appCink}{Qwen3-14B} & 825 & 19.4 & \textbf{67.8} & 99.9 & \textbf{+48.4} \\
\textcolor{appCink}{Qwen3-32B} & 371 & 32.6 & \textbf{62.3} & 88.9 & \textbf{+29.6} \\
\textcolor{appCink}{Qwen3-235B} & 522 & 32.6 & \textbf{76.2} & 96.0 & \textbf{+43.7} \\
\textcolor{appCink}{Qwen3.6-27B} & 431 & 43.2 & \textbf{81.4} & 91.4 & \textbf{+38.3} \\
\textcolor{appCink}{Qwen3.8-27B} & 384 & 47.1 & \textbf{82.8} & 92.4 & \textbf{+35.7} \\
\textcolor{appCink}{DeepSeek-V4-F} & 258 & 37.2 & \textbf{69.4} & 80.2 & \textbf{+32.2} \\
\textcolor{appCink}{Kimi-K2.5} & 166 & 45.8 & \textbf{79.5} & 86.1 & \textbf{+33.7} \\
\textcolor{appCink}{Kimi-K2.6} & 178 & 45.5 & \textbf{77.0} & 83.1 & \textbf{+31.5} \\
\midrule
\multicolumn{6}{l}{\textcolor{appCink}{\textbf{Paired changes and effect on the complete benchmark}}} \\
\cmidrule(lr){1-6}\addlinespace[3pt]
\textbf{Model} & \shortstack{\textbf{Errors}\\\textbf{corrected (\%)}} & \shortstack{\textbf{Correct diagnoses}\\\textbf{lost (\%)}} & \multicolumn{2}{c}{\shortstack{\textbf{95\% CI for accuracy gain}\\\textbf{on selected cases (pp)}}} & \shortstack{\textbf{Gain over all}\\\textbf{1,050 cases (pp)}} \\
\midrule
\textcolor{appCink}{Qwen3-8B} & 47.7 & 2.5 & \multicolumn{2}{c}{\textcolor{appCmuted}{[+42.0, +48.6]}} & +44.9 \\
\textcolor{appCink}{Qwen3-14B} & 50.5 & 2.2 & \multicolumn{2}{c}{\textcolor{appCmuted}{[+44.7, +52.2]}} & +38.0 \\
\textcolor{appCink}{Qwen3-32B} & 32.9 & 3.2 & \multicolumn{2}{c}{\textcolor{appCmuted}{[+24.3, +35.0]}} & +10.5 \\
\textcolor{appCink}{Qwen3-235B} & 45.6 & 1.9 & \multicolumn{2}{c}{\textcolor{appCmuted}{[+39.1, +48.3]}} & +21.7 \\
\textcolor{appCink}{Qwen3.6-27B} & 41.3 & 3.0 & \multicolumn{2}{c}{\textcolor{appCmuted}{[+33.2, +43.6]}} & +15.7 \\
\textcolor{appCink}{Qwen3.8-27B} & 38.8 & 3.1 & \multicolumn{2}{c}{\textcolor{appCmuted}{[+30.2, +40.9]}} & +13.0 \\
\textcolor{appCink}{DeepSeek-V4-F} & 38.4 & 6.2 & \multicolumn{2}{c}{\textcolor{appCmuted}{[+24.8, +39.1]}} & +7.9 \\
\textcolor{appCink}{Kimi-K2.5} & 39.2 & 5.4 & \multicolumn{2}{c}{\textcolor{appCmuted}{[+25.3, +42.8]}} & +5.3 \\
\textcolor{appCink}{Kimi-K2.6} & 37.1 & 5.6 & \multicolumn{2}{c}{\textcolor{appCmuted}{[+23.0, +39.9]}} & +5.3 \\
\bottomrule
\end{tabular}
}
\end{table}

\paragraph{Final diagnosis outcomes.}
Gated accuracy exceeds original accuracy for every LLM in Table~\ref{tab:impact-gate-full}. Among selected cases, 32.9--50.5\% change from an incorrect original diagnosis to a correct diagnosis under the gate, while only 1.9--6.2\% lose an originally correct diagnosis. The resulting net accuracy gains range from 29.6 to 48.4 percentage points, with all nine paired confidence intervals above zero. These are improvements in actual submitted diagnoses, complementing the candidate-belief measurements in Appendix~\ref{app:impact-boundary-rescue}. The gains remain positive after accounting for both newly incorrect diagnoses and cases that end without a submission.

\noindent\begin{minipage}{\linewidth}
The gated submission rate ranges from 80.2\% to 100\%. For example, Qwen3-32B originally submits prematurely on 371 cases, with 32.6\% accuracy on those cases. Under the gate, it submits a diagnosis on 88.9\% of the same cases and achieves 62.3\% accuracy when all 371 cases remain in the denominator. Errors corrected account for 32.9\% of the selected cases and correct diagnoses lost account for 3.2\%, producing a net gain of 29.6 percentage points before rounding. This example illustrates why submission rate and accuracy are reported separately.
\end{minipage}

\paragraph{Effect on the complete benchmark.}
The last column expresses the same paired changes relative to all 1,050 cases. It depends on both how frequently a model submits prematurely and how much the gate improves those cases. For Qwen3-8B, 1,041 of 1,050 cases receive the intervention, so its 45.2-point gain on selected cases corresponds to a 44.9-point gain over the full benchmark. For Kimi-K2.5, 166 cases receive the intervention, so a 33.7-point gain on selected cases corresponds to a 5.3-point gain over all cases. The smaller benchmark-wide gain in the latter example reflects the smaller number of premature submissions as well as the intervention's effect on those submissions.

\FloatBarrier

\section{Additional Evaluation of \METHOD{}}
\label{app:evd-details}

\begingroup
\raggedbottom
\definecolor{appEink}{RGB}{50,78,112}

\subsection{Evaluation Protocol and Diagnosis Outcomes}
We evaluate LLM only and \METHOD{} on \OURS{} with Qwen3.8-27B, DeepSeek-V4-Flash, Qwen3-32B, Kimi-K2.6, and Claude-Opus-4-8, as in Section~\ref{sec:evd-evaluation}. Both methods use the same candidates and evidence order for each case. LLM only chooses when to submit a diagnosis or wait. The harness uses the same frozen model to propose a diagnosis, retrieves CDW knowledge, and checks the proposal against the evidence through \textit{Evidence Witness} before \textit{Submission Controller} permits submission. A rejected proposal becomes a wait, so the next evidence unit can be considered. A trajectory ends at its first submitted diagnosis or when the available evidence is exhausted. Cases ending without a diagnosis count as incorrect in both conditions.

Accuracy is the percentage of all cases with a correct submitted diagnosis. Premature incorrect submission requires both $\tau_i<t_i^*$ and $\hat y_i\ne y_i^*$, while boundary submission requires $\tau_i=t_i^*$ regardless of diagnosis correctness. All rates use the full evaluated set as their denominator, including cases without an answer. Appendix~\ref{app:metric-outcomes} defines the remaining outcome and timing metrics.

\begin{table}[!ht]
\centering\small
\setlength{\tabcolsep}{5pt}
\renewcommand{\arraystretch}{1.13}
\caption{Diagnosis outcomes on \OURS{}. Premature incorrect submissions combine a timing error with a diagnostic error. Boundary submissions occur exactly at $t_i^*$. Bold values indicate the better result within each model comparison.}
\label{tab:evd-additional-outcomes}
\resizebox{\linewidth}{!}{%
\begin{tabular}{@{}lrrrrrr@{}}
\toprule
& \multicolumn{2}{c}{Accuracy $\uparrow$} & \multicolumn{2}{c}{Premature incorrect $\downarrow$} & \multicolumn{2}{c}{Boundary submissions $\uparrow$}\\
\cmidrule(lr){2-3}\cmidrule(lr){4-5}\cmidrule(l){6-7}
Model & LLM only & \METHOD{} & LLM only & \METHOD{} & LLM only & \METHOD{}\\
\midrule
\textcolor{appEink}{Qwen3.8-27B} & 60.9 & \textbf{75.6} & 32.0 & \textbf{4.9} & 41.6 & \textbf{73.8}\\
\textcolor{appEink}{DeepSeek-V4-Flash} & 42.9 & \textbf{68.4} & 14.7 & \textbf{8.9} & 27.2 & \textbf{70.4}\\
\textcolor{appEink}{Qwen3-32B} & 50.9 & \textbf{62.9} & 24.0 & \textbf{7.8} & 38.9 & \textbf{67.1}\\
\textcolor{appEink}{Kimi-K2.6} & 60.2 & \textbf{78.0} & 31.1 & \textbf{7.6} & 36.9 & \textbf{66.4}\\
\textcolor{appEink}{Claude-Opus-4-8} & 34.2 & \textbf{85.3} & 4.9 & \textbf{3.3} & 24.7 & \textbf{68.4}\\
\bottomrule
\end{tabular}%
}
\end{table}

Table~\ref{tab:evd-additional-outcomes} shows that \METHOD{} improves diagnostic accuracy and reduces premature incorrect submissions for all five models. At the same time, more diagnoses are submitted at $t_i^*$, when the available evidence first distinguishes the correct candidate from its alternatives. The main results also show fewer unanswered cases (Table~\ref{tab:evc-outcomes-visual}), so more trajectories end with a submitted diagnosis. Together, these results show improvements in both diagnostic success and the alignment of diagnosis submission with evidence sufficiency.

\begin{table}[htbp]
\centering\small
\setlength{\tabcolsep}{5pt}
\renewcommand{\arraystretch}{1.13}
\caption{Submission timing and accuracy with a $\pm1$-turn boundary tolerance (\%). Correct within window counts correct diagnoses submitted inside $[t^*-1,t^*+1]$. Bold marks the better result per model.}
\label{tab:evd-tolerance-1}
\resizebox{\linewidth}{!}{%
\begin{tabular}{@{}llrrrrrr@{}}
\toprule
\multirow{2}{*}{Model} & \multirow{2}{*}{Method} & \multicolumn{4}{c}{Submission timing} & \multicolumn{2}{c}{Diagnostic outcome}\\
\cmidrule(lr){3-6}\cmidrule(l){7-8}
& & Premature & Within window & Delayed & No answer & Accuracy & Correct within window\\
\midrule
\textcolor{appEink}{Qwen3.8-27B} & LLM only & 41.8 & 53.1 & \textbf{2.4} & 2.7 & 60.9 & 45.1\\
 & \METHOD{} & \textbf{9.8} & \textbf{84.2} & 3.6 & \textbf{2.4} & \textbf{75.6} & \textbf{67.1}\\
\addlinespace
\textcolor{appEink}{DeepSeek-V4-Flash} & LLM only & 17.9 & 37.9 & 4.9 & 39.3 & 42.9 & 33.5\\
 & \METHOD{} & \textbf{13.1} & \textbf{82.4} & \textbf{3.1} & \textbf{1.3} & \textbf{68.4} & \textbf{57.9}\\
\addlinespace
\textcolor{appEink}{Qwen3-32B} & LLM only & 29.1 & 48.7 & \textbf{5.8} & 16.4 & 50.9 & 38.9\\
 & \METHOD{} & \textbf{10.0} & \textbf{78.2} & \textbf{5.8} & \textbf{6.0} & \textbf{62.9} & \textbf{56.7}\\
\addlinespace
\textcolor{appEink}{Kimi-K2.6} & LLM only & 40.7 & 49.3 & \textbf{4.7} & 5.3 & 60.2 & 42.7\\
 & \METHOD{} & \textbf{14.7} & \textbf{75.6} & 6.9 & \textbf{2.9} & \textbf{78.0} & \textbf{63.3}\\
\addlinespace
\textcolor{appEink}{Claude-Opus-4-8} & LLM only & 9.3 & 29.3 & \textbf{1.6} & 59.8 & 34.2 & 27.8\\
 & \METHOD{} & \textbf{6.7} & \textbf{79.3} & 5.6 & \textbf{8.4} & \textbf{85.3} & \textbf{73.1}\\
\bottomrule
\end{tabular}}
\end{table}

\begin{table}[htbp]
\centering\small
\setlength{\tabcolsep}{5pt}
\renewcommand{\arraystretch}{1.13}
\caption{Submission timing and accuracy with a $\pm2$-turn boundary tolerance (\%). Correct within window counts correct diagnoses submitted inside $[t^*-2,t^*+2]$. Bold marks the better result per model.}
\label{tab:evd-tolerance-2}
\resizebox{\linewidth}{!}{%
\begin{tabular}{@{}llrrrrrr@{}}
\toprule
\multirow{2}{*}{Model} & \multirow{2}{*}{Method} & \multicolumn{4}{c}{Submission timing} & \multicolumn{2}{c}{Diagnostic outcome}\\
\cmidrule(lr){3-6}\cmidrule(l){7-8}
& & Premature & Within window & Delayed & No answer & Accuracy & Correct within window\\
\midrule
\textcolor{appEink}{Qwen3.8-27B} & LLM only & 33.8 & 61.3 & 2.2 & 2.7 & 60.9 & 48.4\\
 & \METHOD{} & \textbf{7.8} & \textbf{88.0} & \textbf{1.8} & \textbf{2.4} & \textbf{75.6} & \textbf{69.3}\\
\addlinespace
\textcolor{appEink}{DeepSeek-V4-Flash} & LLM only & 15.2 & 42.0 & 3.6 & 39.3 & 42.9 & 35.3\\
 & \METHOD{} & \textbf{10.7} & \textbf{87.5} & \textbf{0.4} & \textbf{1.3} & \textbf{68.4} & \textbf{62.6}\\
\addlinespace
\textcolor{appEink}{Qwen3-32B} & LLM only & 21.8 & 57.6 & 4.2 & 16.4 & 50.9 & 43.3\\
 & \METHOD{} & \textbf{7.3} & \textbf{84.0} & \textbf{2.7} & \textbf{6.0} & \textbf{62.9} & \textbf{59.3}\\
\addlinespace
\textcolor{appEink}{Kimi-K2.6} & LLM only & 34.4 & 57.3 & \textbf{2.9} & 5.3 & 60.2 & 46.0\\
 & \METHOD{} & \textbf{12.9} & \textbf{80.7} & 3.6 & \textbf{2.9} & \textbf{78.0} & \textbf{67.3}\\
\addlinespace
\textcolor{appEink}{Claude-Opus-4-8} & LLM only & 7.1 & 32.4 & \textbf{0.7} & 59.8 & 34.2 & 30.2\\
 & \METHOD{} & \textbf{3.8} & \textbf{84.9} & 2.9 & \textbf{8.4} & \textbf{85.3} & \textbf{77.3}\\
\bottomrule
\end{tabular}}
\end{table}

\subsection{Sensitivity to Boundary Tolerance}
\label{app:evd-tolerance}
We test whether the timing advantage of \METHOD{} depends on requiring submission at exactly the annotated $t_i^*$. Following Appendix~\ref{app:boundary-tolerance}, we rescore the original trajectories with windows $[t_i^*-k,t_i^*+k]$ for $k=1$ and $k=2$. Submissions before, inside, and after each window are classified as premature, within-window, and delayed. We report non-answers separately.

Tables~\ref{tab:evd-tolerance-1} and~\ref{tab:evd-tolerance-2} show that wider windows classify more submissions as timely under both methods. For each of the five models, \METHOD{} still places substantially more diagnoses near $t_i^*$ and yields more correct diagnoses within the window. Thus, its timing advantage persists when the sufficiency boundary is allowed to vary by one or two turns.

The tolerance analysis reclassifies submission timing without changing any action or final diagnosis. Diagnostic accuracy is therefore unchanged, while the consistent increase in within-window submissions shows that the harness concentrates diagnosis submissions closer to the sufficiency boundary.
\FloatBarrier

\subsection{Separating Knowledge, Verification, and Context Effects}
\label{app:evd-ablations}

\paragraph{Component controls.}
To separate the contributions of CDW knowledge and evidence verification, we compare four named settings on the same 200 \OURS{} cases. Each setting uses the same patient evidence, candidate diagnoses, evidence order, harness execution protocol, and two-action history window. The \textit{Full} setting supplies CDW knowledge to \textit{Answer Proposal} and \textit{Evidence Witness}. The \textit{Submission Controller} then follows the witness verdict to submit or wait. The \textit{$-$CDW} setting removes retrieved knowledge while retaining verification and submission control. The \textit{$-$Witness} setting retains CDW but submits the diagnosis proposed by \textit{Answer Proposal} directly, without calling \textit{Evidence Witness}. The \textit{Both} setting removes CDW and witness review together. The compact header $-$W abbreviates witness removal. Incorrect diagnoses and non-answers count as failures in every setting.

\begin{table}[!htb]
\centering
\definecolor{evdblue}{HTML}{3F6F9F}
\definecolor{evdred}{HTML}{A45760}
\begin{minipage}[t]{0.49\linewidth}
\vspace{0pt}
\caption{Accuracy after removing CDW, Evidence Witness, or both (\%). Red subscripts give differences from Full.}
\label{tab:evd-factor}
\centering\fontsize{7.0}{8.5}\selectfont
\setlength{\tabcolsep}{1.4pt}
\begin{tabular*}{\linewidth}{@{\extracolsep{\fill}}lrrrr@{}}
\toprule
Model & Full & $-$Witness & $-$CDW & Both \\
\midrule
Q3.8-27B & \textbf{\textcolor{evdblue}{75.5}} & 72.0$_{\textcolor{evdred}{-3.5}}$ & 63.0$_{\textcolor{evdred}{-12.5}}$ & 57.5$_{\textcolor{evdred}{-18.0}}$ \\
DS-V4-F & \textbf{\textcolor{evdblue}{68.5}} & 65.0$_{\textcolor{evdred}{-3.5}}$ & 54.0$_{\textcolor{evdred}{-14.5}}$ & 43.0$_{\textcolor{evdred}{-25.5}}$ \\
Q3-32B & \textbf{\textcolor{evdblue}{63.0}} & 58.5$_{\textcolor{evdred}{-4.5}}$ & 50.5$_{\textcolor{evdred}{-12.5}}$ & 46.5$_{\textcolor{evdred}{-16.5}}$ \\
Kimi-K2.6 & \textbf{\textcolor{evdblue}{78.0}} & 74.0$_{\textcolor{evdred}{-4.0}}$ & 66.0$_{\textcolor{evdred}{-12.0}}$ & 60.0$_{\textcolor{evdred}{-18.0}}$ \\
Opus-4-8 & \textbf{\textcolor{evdblue}{86.5}} & 77.5$_{\textcolor{evdred}{-9.0}}$ & 77.5$_{\textcolor{evdred}{-9.0}}$ & 75.0$_{\textcolor{evdred}{-11.5}}$ \\
\bottomrule
\end{tabular*}
\end{minipage}\hfill
\begin{minipage}[t]{0.49\linewidth}
\vspace{0pt}
\caption{Premature incorrect submissions (E) and non-answers (N) under the four component settings (\%).}
\label{tab:evd-factor-safety}
\centering\fontsize{6.0}{8}\selectfont
\setlength{\tabcolsep}{1pt}
\begin{tabular*}{\linewidth}{@{\extracolsep{\fill}}l*{8}{r}@{}}
\toprule
& \multicolumn{2}{c}{Full} & \multicolumn{2}{c}{$-$W} & \multicolumn{2}{c}{$-$CDW} & \multicolumn{2}{c}{Both} \\
\cmidrule(lr){2-3}\cmidrule(lr){4-5}\cmidrule(lr){6-7}\cmidrule(l){8-9}
Model & E & N & E & N & E & N & E & N \\
\midrule
Q3.8-27B & \textbf{3.5} & 2.5 & 8.0 & \textbf{0.5} & 26.0 & 6.5 & 34.0 & 2.5 \\
DS-V4-F & \textbf{6.5} & 1.5 & 12.5 & \textbf{0.0} & 39.5 & 2.5 & 51.5 & 1.0 \\
Q3-32B & \textbf{6.0} & 6.0 & 14.0 & \textbf{0.0} & 29.0 & 11.0 & 36.0 & 4.5 \\
Kimi-K2.6 & \textbf{6.0} & 3.0 & 12.5 & \textbf{0.5} & 24.0 & 7.0 & 31.5 & 6.0 \\
Opus-4-8 & 2.5 & 8.5 & 5.5 & \textbf{1.0} & \textbf{0.5} & 22.0 & 11.5 & 12.5 \\
\bottomrule
\end{tabular*}
\end{minipage}
\end{table}

Table~\ref{tab:evd-factor} shows that removing \textit{Evidence Witness} lowers accuracy by 3.5--9.0 percentage points across the five models. Removing CDW while retaining the witness lowers it by 9.0--14.5 points. Table~\ref{tab:evd-factor-safety} shows that witness removal also increases premature incorrect submissions, with the combined removal generally producing larger increases. CDW thus helps the model distinguish candidate diagnoses, while the witness helps determine when a proposed diagnosis is supported for submission. Accuracy counts unanswered cases as failures throughout.

\paragraph{Retrieval source and contrastive organization.}
We isolate the value of contrastive organization by changing the Wiki pages supplied to \textit{Answer Proposal} and \textit{Evidence Witness}. As described in Section~\ref{sec:evd-offline}, the pages are compiled from independent case reports after excluding evaluation sources. \textit{None} supplies no pages. The other controls retrieve disease pages, which summarize source-supported clinical findings. \textit{Random} samples three pages at the first evidence turn; \textit{BM25} selects three by text similarity to \begin{table}[!htb]
\centering\small
\definecolor{evdblue}{HTML}{3F6F9F}
\definecolor{evdred}{HTML}{A45760}
\definecolor{evdgreen}{HTML}{477F79}
\definecolor{evdgray}{HTML}{687482}
\setlength{\tabcolsep}{5pt}\renewcommand{\arraystretch}{1.13}
\caption{Accuracy and premature incorrect submissions under five knowledge views (\%). Bold marks the best result for each model and metric.}
\label{tab:evd-retrieval-full}
\vspace{0.3em}\resizebox{\linewidth}{!}{%
\begin{tabular}{@{}lrrrrrrrrrr@{}}
\toprule
& \multicolumn{5}{c}{Accuracy $\uparrow$} & \multicolumn{5}{c}{Premature incorrect $\downarrow$} \\
\cmidrule(lr){2-6}\cmidrule(l){7-11}
Model & None & Random & BM25 & Contrast. & CDW & None & Random & BM25 & Contrast. & CDW \\
\midrule
\textcolor{appEink}{Qwen3.8-27B} & 63.0 & 64.0 & 67.0 & 72.0 & \textbf{75.5} & 26.0 & 12.5 & 9.0 & 5.0 & \textbf{3.5} \\
\textcolor{appEink}{DeepSeek-V4-Flash} & 54.0 & 55.0 & 59.0 & 65.0 & \textbf{68.5} & 39.5 & 17.0 & 11.5 & 9.0 & \textbf{6.5} \\
\textcolor{appEink}{Qwen3-32B} & 50.5 & 51.5 & 54.5 & 59.5 & \textbf{63.0} & 29.0 & 17.5 & 11.5 & 7.5 & \textbf{6.0} \\
\textcolor{appEink}{Kimi-K2.6} & 66.0 & 67.0 & 70.0 & 75.0 & \textbf{78.0} & 24.0 & 17.5 & 13.0 & 7.5 & \textbf{6.0} \\
\textcolor{appEink}{Claude-Opus-4-8} & 77.5 & 78.5 & 80.5 & 84.5 & \textbf{86.5} & \textbf{0.5} & 1.5 & 1.0 & \textbf{0.5} & 2.5 \\
\bottomrule
\end{tabular}}
\end{table}
that evidence and the candidates. \textit{Contrastive} organizes the same BM25 pages by candidate pair in $\mathcal{Y}_i$, highlighting shared and distinguishing findings. \textit{Full CDW} updates retrieval as evidence arrives and follows links from disease to test and bridge pages, connecting findings to distinctions among candidates.

Table~\ref{tab:evd-retrieval-full} shows that the \textit{Full CDW} setting achieves the highest diagnostic accuracy across all five models. Starting from the same BM25 page retrieval, \textit{Contrastive} raises accuracy by 4.0--6.0 percentage points and reduces premature incorrect submissions for each model. This comparison shows the value of organizing retrieved pages by the distinctions among the current candidate diagnoses, beyond matching pages to the patient text. The linked pages of the complete Wiki provide a richer view of those distinctions and support the strongest overall accuracy.

\paragraph{Action history and the execution protocol.}
In addition to retrieving knowledge and verifying diagnoses, the harness changes how many of its earlier wait and diagnosis-submission actions the model can see. We examine whether this shorter action record affects diagnostic outcomes and whether the harness execution procedure contributes separately. The \textit{Natural} setting is the original LLM-only procedure with the complete patient record and prior actions. \textit{Short} uses the same prompt and response processing and retains all patient evidence, but shows only the two most recent actions. \textit{Direct} also shows two actions, while using the harness response format, parsing, and state updates to submit diagnosis proposals directly. None of these settings retrieves CDW knowledge or calls \textit{Evidence Witness}, so the two successive comparisons examine the visible action record and execution procedure in turn.

\begin{table}[!htb]
\centering\small
\definecolor{evdblue}{HTML}{3F6F9F}
\definecolor{evdred}{HTML}{A45760}
\definecolor{evdgreen}{HTML}{477F79}
\definecolor{evdgray}{HTML}{687482}
\setlength{\tabcolsep}{5pt}\renewcommand{\arraystretch}{1.13}
\caption{Accuracy and non-answer rates (\%) under history and execution controls. Small numbers show changes from the preceding condition in percentage points. Bold marks the best result per model and metric.}
\label{tab:evd-history}
\vspace{0.3em}\resizebox{\linewidth}{!}{%
\begin{tabular}{@{}lrrrrrr@{}}
\toprule
& \multicolumn{3}{c}{Accuracy $\uparrow$} & \multicolumn{3}{c}{No answer $\downarrow$} \\
\cmidrule(lr){2-4}\cmidrule(l){5-7}
Model & Natural & Short & Direct & Natural & Short & Direct \\
\midrule
\textcolor{appEink}{Qwen3.8-27B} & 59.5 & \textbf{63.5}$_{\textcolor{evdgreen}{+4.0}}$ & 57.5$_{\textcolor{evdred}{-6.0}}$ & 3.5 & \textbf{2.5}$_{\textcolor{evdgreen}{-1.0}}$ & \textbf{2.5}$_{\textcolor{evdgray}{+0.0}}$ \\
\textcolor{appEink}{DeepSeek-V4-Flash} & 45.5 & \textbf{49.5}$_{\textcolor{evdgreen}{+4.0}}$ & 43.0$_{\textcolor{evdred}{-6.5}}$ & 39.5 & \textbf{1.0}$_{\textcolor{evdgreen}{-38.5}}$ & \textbf{1.0}$_{\textcolor{evdgray}{+0.0}}$ \\
\textcolor{appEink}{Qwen3-32B} & 54.0 & \textbf{58.0}$_{\textcolor{evdgreen}{+4.0}}$ & 46.5$_{\textcolor{evdred}{-11.5}}$ & 14.5 & \textbf{4.0}$_{\textcolor{evdgreen}{-10.5}}$ & 4.5$_{\textcolor{evdred}{+0.5}}$ \\
\textcolor{appEink}{Kimi-K2.6} & 61.5 & \textbf{65.5}$_{\textcolor{evdgreen}{+4.0}}$ & 60.0$_{\textcolor{evdred}{-5.5}}$ & 5.0 & \textbf{3.5}$_{\textcolor{evdgreen}{-1.5}}$ & 6.0$_{\textcolor{evdred}{+2.5}}$ \\
\textcolor{appEink}{Claude-Opus-4-8} & 37.5 & 46.5$_{\textcolor{evdgreen}{+9.0}}$ & \textbf{75.0}$_{\textcolor{evdgreen}{+28.5}}$ & 56.0 & 36.0$_{\textcolor{evdgreen}{-20.0}}$ & \textbf{12.5}$_{\textcolor{evdgreen}{-23.5}}$ \\
\bottomrule
\end{tabular}}
\end{table}

Table~\ref{tab:evd-history} shows accuracy increases of 4.0--9.0 points in the observed results after action history is shortened, alongside generally fewer non-answers. The Direct condition additionally changes response formatting and state handling, showing that the execution procedure also affects diagnostic outcomes. Accordingly, the component and retrieval comparisons above hold both the history window and execution protocol fixed when assessing CDW and \textit{Evidence Witness}.

\subsection{Sensitivity to Evidence Perturbations}
We evaluate robustness on \OURS{} using the evidence interventions associated with \hyperref[rq:order]{RQ3} and \hyperref[rq:strength]{RQ4}, with implementation details in Appendix~\ref{app:mechanism-details}. Each case is run with its original evidence sequence and with the corresponding perturbed sequence. The ground-truth diagnosis and candidate set remain fixed. Evidence reordering changes the presentation order of the same clinical information. The insertion conditions add repeated evidence, irrelevant text, or a paraphrase of existing evidence. Repeated and paraphrased evidence restate information already present, while irrelevant text adds no information that distinguishes the diagnostic candidates.

For these experiments, we compare the final outputs of the original and perturbed diagnosis trajectories using the diagnosis outcome disagreement defined in Appendix~\ref{app:metric-interventions}. A flip occurs when the submitted diagnoses differ or when only one trajectory submits a diagnosis. Two unanswered trajectories count as agreement. Table~\ref{tab:evd-insertions} compares LLM only and \METHOD{} under each insertion condition and reports their means across the three conditions. Results for evidence reordering appear in the \emph{Order flip} column of Table~\ref{tab:evc-outcomes-visual}.

\begin{table}[!ht]
\centering\small
\setlength{\tabcolsep}{5pt}
\renewcommand{\arraystretch}{1.13}
\caption{Answer-flip rates under evidence insertions (\%, lower is better). Each method is evaluated under the same three perturbations. Mean denotes the average across conditions. Bold marks the lower rate in each paired comparison.}
\label{tab:evd-insertions}
\resizebox{\linewidth}{!}{%
\begin{tabular}{@{}lrrrrrrrr@{}}
\toprule
& \multicolumn{4}{c}{LLM only} & \multicolumn{4}{c}{\METHOD{}}\\
\cmidrule(lr){2-5}\cmidrule(l){6-9}
Model & \shortstack{Repeated\\evidence} & \shortstack{Irrelevant\\text} & \shortstack{Paraphrased\\evidence} & Mean & \shortstack{Repeated\\evidence} & \shortstack{Irrelevant\\text} & \shortstack{Paraphrased\\evidence} & Mean\\
\midrule
\textcolor{appEink}{Qwen3.8-27B} & 9.7 & 8.5 & 8.3 & 8.8 & \textbf{1.6} & \textbf{1.8} & \textbf{1.6} & \textbf{1.7}\\
\textcolor{appEink}{DeepSeek-V4-Flash} & 31.7 & 27.0 & 29.0 & 29.2 & \textbf{1.1} & \textbf{1.1} & \textbf{1.3} & \textbf{1.2}\\
\textcolor{appEink}{Qwen3-32B} & 13.3 & 13.3 & 11.9 & 12.8 & \textbf{3.4} & \textbf{3.8} & \textbf{2.7} & \textbf{3.3}\\
\textcolor{appEink}{Kimi-K2.6} & 21.6 & 18.7 & 20.7 & 20.3 & \textbf{1.8} & \textbf{1.1} & \textbf{1.6} & \textbf{1.5}\\
\textcolor{appEink}{Claude-Opus-4-8} & 30.8 & 30.6 & 31.5 & 31.0 & \textbf{3.4} & \textbf{3.1} & \textbf{3.4} & \textbf{3.3}\\
\bottomrule
\end{tabular}%
}
\end{table}

\METHOD{} consistently reduces answer flips across repeated evidence, irrelevant text, and paraphrased evidence (Table~\ref{tab:evd-insertions}). Rates fall from 8.3--31.7\% with LLM only to 1.1--3.8\% with the harness. Averaged across the three conditions, the reduction ranges from 74\% to 96\% relative to LLM only. The improvement therefore extends across both repetition and changes in wording, supporting greater stability of diagnosis outputs when additional text supplies no new diagnostic information.

We also evaluate misleading evidence using the 318 paired cases described in \hyperref[rq:strength]{RQ4}. Each pair varies the strength of evidence favoring a designated incorrect diagnosis while retaining the underlying case and subsequent corrective evidence. The \emph{Target error} column of Table~\ref{tab:evc-outcomes-visual} reports submission of that incorrect diagnosis before corrective evidence arrives in the high-strength condition. \METHOD{} reduces this error rate relative to LLM only for all five models. This result complements the insertion experiments by showing improved resistance to evidence that actively favors a wrong diagnosis, in addition to reduced sensitivity to redundant or irrelevant text.

\subsection{Transfer to Static Medical Question Answering}
\label{app:evd-static-qa}
Static medical question answering tests whether the benefit of \METHOD{} extends beyond deciding when to answer during progressive evidence disclosure. In this setting, all question information is available from the outset. Improvements must therefore come from how the model uses that information to select and verify a diagnosis. We evaluate the same five models on DiagnosisArena-MCQ and MedXpertQA-Text, with results in Table~\ref{tab:evd-ood}.

\begin{table}[!htb]
\centering\small
\setlength{\tabcolsep}{5mm}
\renewcommand{\arraystretch}{1.13}
\caption{DiagnosisArena-MCQ accuracy (\%). Direct sampling aggregates the indicated number of independently generated answers. The single-sample column is LLM only. Bold marks the highest accuracy for each model.}
\label{tab:evd-qa-controls}
\resizebox{\linewidth}{!}{%
\begin{tabular}{@{}lrrrrrr@{}}
\toprule
& \multicolumn{3}{c}{Direct sampling} & \multicolumn{3}{c}{\METHOD{}}\\
\cmidrule(lr){2-4}\cmidrule(l){5-7}
Model & 1 sample & 5 samples & 6 samples & Full & $-$Witness & $-$CDW\\
\midrule
\textcolor{appEink}{Qwen3.8-27B} & 54.1 & 57.8 & 58.0 & 57.1 & \textbf{64.8} & 53.9\\
\textcolor{appEink}{DeepSeek-V4-Flash} & 52.0 & 53.5 & 53.6 & 56.1 & \textbf{59.9} & 53.5\\
\textcolor{appEink}{Qwen3-32B} & 50.2 & 50.8 & 51.7 & 49.6 & \textbf{51.8} & 46.2\\
\textcolor{appEink}{Kimi-K2.6} & 63.7 & 64.0 & 64.9 & 67.5 & \textbf{72.6} & 64.3\\
\textcolor{appEink}{Claude-Opus-4-8} & 67.8 & 67.4 & 67.3 & 69.8 & \textbf{77.6} & 63.3\\
\bottomrule
\end{tabular}%
}
\end{table}

Each model receives the complete question and its answer candidates in a single turn. The LLM-only baseline directly selects an answer. \METHOD{} treats the question as the available patient record and applies CDW retrieval, Differential Guidance, Answer Proposal, and Evidence Witness before diagnosis submission. The witness uses the same question information to assess the proposal, and its citations are checked against that information. A question with no authorized answer counts as incorrect. Accuracy therefore includes all questions for both methods. Since no further clinical evidence is disclosed, this evaluation uses accuracy without assigning a sufficiency boundary or measuring premature submissions.

\begin{table}[!htb]
\centering\small
\definecolor{evdblue}{HTML}{3F6F9F}
\definecolor{evdred}{HTML}{A45760}
\definecolor{evdgreen}{HTML}{477F79}
\definecolor{evdgray}{HTML}{687482}
\setlength{\tabcolsep}{7mm}\renewcommand{\arraystretch}{1.13}
\caption{All-sample accuracy (\%) on DiagnosisArena-MCQ and MedXpertQA-Text. Unanswered questions count as errors. Bold marks the better method for each model and benchmark.}
\label{tab:evd-ood}
\vspace{0.3em}\resizebox{\linewidth}{!}{%
\begin{tabular}{@{}lrrrr@{}}
\toprule
& \multicolumn{2}{c}{DiagnosisArena-MCQ} & \multicolumn{2}{c}{MedXpertQA-Text} \\
\cmidrule(lr){2-3}\cmidrule(lr){4-5}
Model & LLM only & \METHOD{} & LLM only & \METHOD{} \\
\midrule
\textcolor{appEink}{Qwen3.8-27B} & 54.1 & \textbf{57.1} & 24.4 & \textbf{27.9} \\
\textcolor{appEink}{DeepSeek-V4-Flash} & 52.0 & \textbf{56.1} & 15.1 & \textbf{22.1} \\
\textcolor{appEink}{Qwen3-32B} & \textbf{50.2} & 49.6 & 10.5 & \textbf{18.6} \\
\textcolor{appEink}{Kimi-K2.6} & 63.7 & \textbf{67.5} & 17.4 & \textbf{20.9} \\
\textcolor{appEink}{Claude-Opus-4-8} & 67.8 & \textbf{69.8} & 38.4 & \textbf{44.2} \\
\bottomrule
\end{tabular}}
\end{table}

To test whether gains on static QA arise from diagnostic knowledge, verification, or simply generating more answers, we compare a single direct answer, five or six independent answer samples, Full \METHOD{}, and two component removals on DiagnosisArena-MCQ (Table~\ref{tab:evd-qa-controls}). Direct sampling selects the most frequent label from the same complete question. Ties follow a fixed label order, and invalid labels result in an unanswered, incorrect item. The $-$Witness condition retains CDW retrieval and diagnosis proposals but submits each proposal without witness review. The $-$CDW condition retains witness review but removes retrieved knowledge. All conditions receive the complete question at once.

Table~\ref{tab:evd-qa-controls} shows that CDW-guided diagnosis proposals without witness review attain the highest DiagnosisArena-MCQ accuracy for each model. Removing CDW lowers the Full result, demonstrating the value of retrieved differential knowledge when the entire question is available at once. On MedXpertQA-Text, Full \METHOD{} improves over LLM only for all five models (Table~\ref{tab:evd-ood}). These results show that CDW-supported diagnosis also transfers to static medical QA.

\clearpage
\endgroup

\end{document}